\RequirePackage{fix-cm}

\documentclass[twocolumn]{svjour3}          
\smartqed  
\usepackage{graphicx}
\usepackage{url}
\usepackage{amsmath}
\usepackage{multirow}
\usepackage{amssymb}
\usepackage{bm}
\usepackage{bbm}
\usepackage{booktabs}
\usepackage[table]{xcolor}
\definecolor{mydarkgreen}{RGB}{0,100,0} 

\usepackage[colorlinks,
            linkcolor=red,
            anchorcolor=blue,
            citecolor=blue,
            CJKbookmarks=true,
            breaklinks=true]{hyperref}

\usepackage[ruled,vlined]{algorithm2e}

\makeatletter
\@fleqnfalse
\makeatother
\begin{document}

\title{Generalized Fine-Grained Category Discovery with Multi-Granularity Conceptual Experts
}


\author{Haiyang Zheng         \and
        Nan Pu \and
        Wenjing Li \and
        Nicu Sebe \and
        Zhun Zhong
}


\institute{H. Zheng, N. Pu, and N. Sebe \at Department of Information Engineering and Computer Science, University of Trento, Trento, Italy. 
              \email{\{haiyang.zheng, nan.pu, niculae.sebe\}@unitn.it}           
           \and
           Wenjing Li and Z. Zhong \at
              School of Computer Science and Information Engineering, Hefei University of Technology, China.
              \email{wenjingli@hfut.edu.cn, zhunzhong007@gmail.com}
}

\date{Received: date / Accepted: date}

\maketitle

\begin{abstract}
Generalized Category Discovery (GCD) is an open-world problem that clusters unlabeled data by leveraging knowledge from partially labeled categories. 
A key challenge is that unlabeled data may contain both known and novel categories. Existing approaches suffer from two main limitations. First, they fail to exploit multi-granularity conceptual information in visual data, which limits representation quality. Second, most assume that the number of unlabeled categories is known during training, which is impractical in real-world scenarios. 
To address these issues, we propose a Multi-Granularity Conceptual Experts (MGCE) framework that adaptively mines visual concepts and integrates multi-granularity knowledge for accurate category discovery. 
MGCE consists of two modules: (1) Dynamic Conceptual Contrastive Learning (DCCL), which alternates between concept mining and dual-level representation learning to jointly optimize feature learning and category discovery; and (2) Multi-Granularity Experts Collaborative Learning (MECL), which extends the single-expert paradigm by introducing additional experts at different granularities and by employing a concept alignment matrix for effective cross-expert collaboration. 
Importantly, MGCE can automatically estimate the number of categories in unlabeled data, making it suitable for practical open-world settings. 
Extensive experiments on nine fine-grained visual recognition benchmarks demonstrate that MGCE achieves state-of-the-art results, particularly in novel-class accuracy. Notably, even without prior knowledge of category numbers, MGCE outperforms parametric approaches that require knowing the exact number of categories, with an average improvement of 3.6\%. Code is available at \url{https://github.com/HaiyangZheng/MGCE}.
\end{abstract}

\section{Introduction}
\label{intro}
Deep learning has achieved remarkable success in visual recognition, often outperforming humans on specific benchmarks, but this success typically reflects the strength of specialized models over generalist human cognition. Current vision models are trained on large, labeled datasets within a predefined category set, excelling at recognizing instances of seen classes but struggling with novel categories. Although techniques like open-set recognition~\cite{geng2020recent,anomaly_dection,oodd} attempt to filter out unknown category instances, simply rejecting them is insufficient. In contrast, human visual recognition is more flexible, as humans are able not only to identify novel categories but can also group related instances together based on shared features, even without prior knowledge of these novel categories. The cognitive ability to generalize and infer novel categories presents an ongoing challenge for machine learning models.

\begin{figure}[!t]
  \begin{center}
\includegraphics[width=0.49\textwidth]{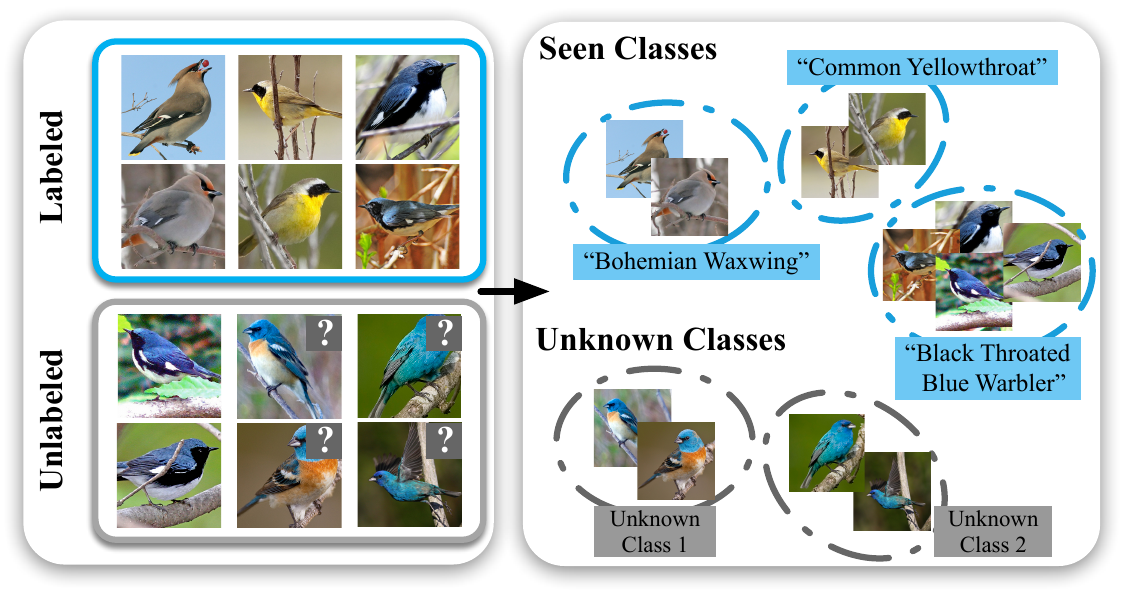}
\caption{Pipeline of Generalized Category Discovery (GCD). GCD aims to exploit knowledge from labeled data to automatically cluster unlabeled data, which contain both known and unknown categories.}
\label{fig:setting}
  \end{center}
  \vspace{-0.1cm}
\end{figure}

\par
To tackle this challenge, the rapidly growing research area of Novel Category Discovery (NCD)~\cite{ncd,ncl,uno,dualrs,openmix,incd,ncdss,rankstats}, has attracted significant attention. In the NCD setting, given a labeled support dataset consisting of known categories and an unlabeled dataset containing unknown categories, the goal is to identify unknown categories by clustering the query set into several groups, where each group shares the same latent category. Recently, Generalized Category Discovery (GCD)~\cite{gcd} has extended the NCD task, as shown in Fig.~\ref{fig:setting}, by assuming that the unlabeled dataset may contain both known and novel classes. This makes GCD a more practical yet challenging problem. 

\begin{figure*}[t]
  \begin{center}
\includegraphics[width=\textwidth]{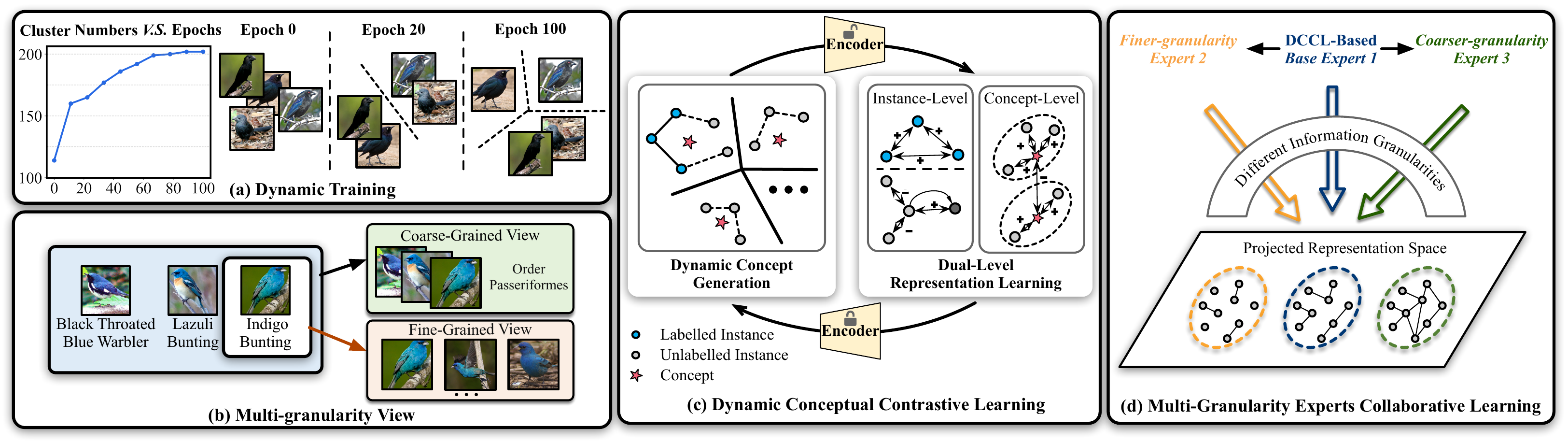}
\caption{(a) Schematic illustration of the proposed dynamic category discovery process.
(b) Schematic illustration of perspectives at different information granularities. 
(c) Framework of the proposed Dynamic Conceptual Contrastive Learning (DCCL) module, which incorporates Dynamic Concept Generation and Dual-Level Representation Learning.  
(d) Framework of the proposed Multi-Granularity Experts Collaborative Learning (MECL) module, which introduces multiple views at different levels of information granularity.}
\label{fig:highlight}
  \end{center}
  \vspace{-0.5cm}
\end{figure*}

\par
To address this problem, recent works propose numerous GCD frameworks~\cite{gcd,dccl,promptcal,gpc,simgcd,pim,ugcd,cms,sptnet,legogcd}, which can be broadly grouped into \textit{parametric}~\cite{simgcd,ugcd,sptnet,legogcd} and \textit{non-parametric} methods~\cite{gcd,promptcal,gpc,pim,cms}. The former paradigm relies on a parametric classifier to infer category labels, whereas the latter leverages offline clustering algorithms to assign cluster indices as category labels. Although both paradigms have demonstrated promising performance on most GCD benchmarks, they overlook the underlying inter-instance relationships across multi-granularity concepts (\textit{e.g.}, target-class, super-class, and sub-class) during training. Due to a lack of rich contextual information from multi-granularity perspectives, these methods fail to effectively learn a discriminative representation, especially for fine-grained visual recognition tasks where inter-class variations are often subtle. Moreover, parametric methods~\cite{simgcd,ugcd,sptnet,legogcd} commonly assume that the number of categories that need to be discovered is known during training, which is not practical in real-world GCD applications. To overcome these limitations, we introduce a novel framework, \textbf{Multi-Granularity Conceptual Experts (MGCE)}, which leverages multi-granularity conceptual knowledge to learn more discriminative GCD representations and can automatically estimate the number of categories without requiring prior knowledge during training. 
MGCE is composed of two complementary modules: \textbf{Dynamic Conceptual Contrastive Learning (DCCL)} and \textbf{Multi-Granularity Experts Collaborative Learning (M-ECL)}. 
DCCL alternates between concept generation and contrastive learning under a single-expert setting across different training epochs, while MECL integrates multiple experts operating at distinct granularities to enable collaborative learning within each epoch.

\par
In DCCL, the goal is to dynamically extract conceptual knowledge from both labeled and unlabeled data at different training stages and leverage it to enhance representation learning. 
As shown in Fig.~\ref{fig:highlight} (a), the number of discovered concepts grows progressively during training. 
In the early epochs, the model tends to cluster visually similar instances into a few shared concepts. 
With increasingly discriminative features, these coarse concepts are gradually partitioned into more fine-grained ones. 
Motivated by this observation, DCCL alternates between two complementary components: Dynamic
Concept Generation (DCG) and Dual-Level Representation Learning (DRL). 
DCG adaptively generates concepts via a semi-supervised clustering strategy without requiring the number of categories in advance, while DRL jointly optimizes instance-level and concept-level objectives to refine representations. 
As illustrated in Fig.~\ref{fig:highlight} (c), DCG and DRL are executed in an alternating fashion throughout training, ensuring consistency between category discovery and representation learning. 
Through this iterative process, DCCL integrates discovery with learning, thereby forming a concept-expert branch tailored to the GCD task.

\par
In our previous work~\cite{dccl}, we employed a single concept-expert branch throughout training to enhance GCD performance. 
However, this design fails to simultaneously incorporate multi-granularity concepts, which often provide complementary knowledge within each training epoch, leading to single-view concept generation. 
For example, as illustrated in Fig.~\ref{fig:highlight} (b), from a coarse-grained perspective, Bunting and Warbler both belong to the order Passeriformes, making them more similar than to birds from other orders. 
From a fine-grained perspective, variations of Indigo Bunting (e.g., flying vs. standing postures) offer a more comprehensive understanding of the category. We therefore propose MECL, a natural extension of our earlier work\cite{dccl}, to mine and integrate multi-granularity conceptual knowledge. Specifically, we design a multi-expert paradigm that expands the base expert into three expert branches operating at distinct granularities (Fig.~\ref{fig:highlight} (d)), and introduce a concept alignment matrix to enable collaborative learning among experts. 
Through this design, coarse- and fine-grained experts provide complementary category insights to the base expert (see Sec.~\ref{exp:further_analysis}). 
By synergistically integrating DCCL and MECL, our MGCE framework achieves more accurate category discovery on fine-grained GCD benchmarks.
\textbf{Importantly}, to stabilize training in the early stage, we extend the representation learning strategy of our previous work~\cite{dccl} by introducing a switchable parametric classifier. 
This module enables our framework to seamlessly adapt to both cases where the number of unlabeled categories is either known or unknown, thereby improving the flexibility and practicality of the learning process. In the inference phase, we directly adopt the concept partitions produced by the base expert as the final clustering output. 

\par
In summary, our contributions can be summarized as follows:
\begin{itemize}
    \item We highlight the importance of dynamically mining concept knowledge at multiple granularities from both labeled and unlabeled data to advance GCD.
    
    \item We introduce the Multi-Granularity Conceptual Experts (MGCE) framework, which jointly achieves category discovery and representation learning without requiring prior knowledge of the number of categories.
    
    \item We develop two complementary modules: (i) Dynamic Conceptual Contrastive Learning (DCCL), which adaptively generates concepts across training stages and enables dual-level representation learning; and (ii) Multi-Granularity Experts Collaborative Learning (MECL), which integrates complementary coarse- and fine-grained concepts to enhance discriminability.
    
    \item Extensive experiments on nine fine-grained benchmarks demonstrate that MGCE establishes new state-of-the-art performance. Notably, even without knowing the number of categories, MGCE surpasses the leading parametric method by an average improvement of 3.6\% in All ACC.
\end{itemize}

\section{Related Work}
\subsection{Novel Category Discovery} 
Novel Category Discovery (NCD) aims to cluster previously unseen categories within unlabeled data by leveraging prior knowledge from labeled categories~\cite{ncd,ncl,uno,dualrs,openmix,incd,ncdss,rankstats}. 
Recently, this task has been extended to Generalized Category Discovery (GCD)~\cite{gcd}, where unlabeled data may contain both known and unknown categories. 
Beyond the standard setting, more practical scenarios have been explored, including Federated GCD~\cite{fgcd}, On-the-fly Category Discovery~\cite{ocd,phe,liu2025generate}, Ultra-fine Category Discovery~\cite{ultrafine}, and multimodal extensions~\cite{clipgcd,textgcd,mgcd,get}. 
Existing approaches can be broadly categorized into \textbf{parametric} methods~\cite{simgcd,ugcd,sptnet,legogcd,protogcd}, which rely on a classifier with a predefined number of classes, and \textbf{non-parametric} methods~\cite{gcd,promptcal,gpc,pim,cms}, which avoid such assumptions (see Tab.~\ref{tab:setting}). 
In addition, recent studies~\cite{aplgcd,congcd,hypcd} have proposed plug-and-play modules that can be integrated into existing GCD frameworks and consistently improve category discovery performance.

\begin{table}[th]
    \centering
    \caption{Comparison of existing GCD frameworks. ``Para.'' indicates \emph{Parametric} approaches, ``Non-Para.'' indicates \emph{Nonparametric} approaches, and ``\#Cls'' indicates the number of classes.}
    \label{tab:setting}
    \setlength{\tabcolsep}{0.8pt}
    \renewcommand{\arraystretch}{1}
    \resizebox{0.49\textwidth}{!}{%
        \begin{tabular}{l|c|c|c}
            \toprule
            \multirow{2}{*}{Methods} & Training & Testing & Exploit Multiple \\
            &  $w/o$ \#Cls & $w/o$ \#Cls & Granularity \\
            \midrule
            Para.~\cite{simgcd,ugcd,sptnet,legogcd,protogcd} & \text{\sffamily x} & \text{\sffamily x} & \text{\sffamily x} \\
            \midrule
            Static Non-Para.~\cite{gcd,promptcal,pim} & \checkmark & \text{\sffamily x} & \text{\sffamily x} \\
            Dynamic Non-Para.~\cite{gpc,cms} & \checkmark & \checkmark & \text{\sffamily x} \\
            \midrule
            MGCE (Ours) & \checkmark & \checkmark & \checkmark \\
            \bottomrule
        \end{tabular}%
    }
\end{table}
\par
\textbf{Parametric frameworks} employ classifiers with a predefined number of categories. 
SimGCD~\cite{simgcd} introduced this paradigm by using self-distillation with pseudo-labels and entropy regularization. 
$\mu$GCD~\cite{ugcd} extended SimGCD with a mean-teacher~\cite{meanteacher} strategy, improving performance on multi-criteria synthetic data-sets. 
SPTNet~\cite{sptnet} introduces a two-stage iterative learning framework for GCD, alternating between model fine-tuning and data parameter optimization via prompt learning. 
LegoGCD~\cite{legogcd} further regularized the learning process with local entropy and dual-view KL divergence, alleviating catastrophic forgetting of known classes.

\par
\textbf{Non-parametric frameworks} avoid specifying the number of categories beforehand. 
Vaze et al.~\cite{gcd} introduced the GCD setting by combining self-supervised and supervised contrastive learning with a semi-superv-ised $k$-means inference, a paradigm later extended by many works. 
PromptCAL~\cite{promptcal} leverages prompt vectors and an affinity graph to refine pseudo-positive sample selection, while PIM~\cite{pim} employs weighted mutual information to handle imbalanced category distributions. 
However, such \textbf{static} methods decouple category discovery from representation learning and require a predefined or separately estimated class number. 
More recent \textbf{dynamic} methods, such as GPC~\cite{gpc} and CMS~\cite{cms}, integrate category estimation with representation learning. 
GPC follows an EM-like framework with split-and-merge operations but overlooks hierarchical structures, whereas CMS iteratively updates embeddings and centroids via Mean Shift but depends on a validation set, limiting practicality. 
\textit{Our MGCE differs by jointly performing category estimation and representation learning without external validation, while exploiting multi-granularity insights.}

\subsection{Contrastive Learning Based on Memory Buffers}
\par\noindent
Contrastive learning (CL)~\cite{chen2020simple,wang2021,wang2022} has proven effective for self-supervised representation learning. 
MoCo~\cite{he2020momentum} showed that maintaining an instance-level memory buffer enables effective sampling of positive and negative pairs, alleviating the need for large batch sizes. 
Building on this idea, Prototypical Contrastive Learning (PCL)~\cite{pcl} compares instance features against prototypes rather than all instances, providing stronger supervision but still requiring memory-intensive instance storage. 
More recently, SCL~\cite{scl} introduced cluster-level momentum prototypes to reduce memory cost, but assumes a fixed number of categories during training, which restricts its applicability. 
\textit{In contrast, our MGCE framework dynamically discovers categories at each epoch, enabling a mutual reinforcement loop where improved category discovery facilitates representation learning, and enhanced representations further promote accurate discovery.}

\subsection{Semi-Supervised Learning and Clustering}
\par\noindent
Semi-supervised learning (SSL) has been extensively studied~\cite{semisurvey}. 
Unlike GCD, SSL typically assumes that labeled and unlabeled data share the same set of classes. 
Among SSL approaches, consistency-based methods such as Mean Teacher~\cite{meanteacher}, MixMatch~\cite{mixmatch}, and FixMatch~\cite{fixmatch} have proven particularly effective, enforcing prediction consistency under input perturbations to exploit unlabeled data. Semi-supervised clustering, however, introduces additional challenges, as supervision may appear in diverse forms~\cite{lange2005learning}. 
For example, pairwise constraints (e.g., must-link or cannot-link relations) can be imposed, or partial cluster assignments can be provided. 
Basu et al.~\cite{basu} further proposed initializing clusters using data points with known labels. 
Nonetheless, these approaches generally assume a fixed number of clusters and lack the flexibility required for adaptive concept discovery. 
\textit{To overcome this limitation, we extend the classical Infomap~\cite{infomap} by incorporating both labeled and unlabeled data, enabling dynamic concept generation with an adaptively varying number of clusters.}

\section{Method}
\subsection{Problem Setup and Framework Overview}
\noindent \textbf{Problem Setup.} 
Generalized category discovery (GCD) aims at automatically categorizing unlabeled images containing both known (old) and unknown (novel) classes based on knowledge from labeled images of known classes. Formally, we are given a labeled dataset, denoted as $\mathcal{D}_{L} = \left\{ (\mathbf{x}^{l}_{i}, y^{l}_{i}) \mid (\mathbf{x}^{l}_{i}, y^{l}_{i}) \in \mathcal{X} \times \mathcal{Y}_{L} \right\}_{i=1}^{N}$, and an unlabeled dataset, denoted as $\mathcal{D}_{U} = \left\{ \mathbf{x}^{u}_{i} \mid \mathbf{x}^{u}_{i} \in \mathcal{X} \right\}_{i=1}^{M}$, where $N$ and $M$ indicate the number of samples in the labeled and unlabeled datasets respectively and $\mathcal{Y}_{L}$ indicates the label spaces of the labeled dataset. The underlying label space of the unlabeled dataset $\mathcal{Y}_{U}$ consists of both the known classes from $\mathcal{Y}_{L}$ and the unknown classes, i.e., $\mathcal{Y}_{L} \subset \mathcal{Y}_{U}$.
The complete training dataset is $\mathcal{D} = \mathcal{D}_{L} \cup \mathcal{D}_{U}$. The number of labeled classes $K_{L}$ can be directly calculated from the labeled data, while the number of unlabeled classes $K_{U}$ is likely unknown during model training. This represents a more realistic open-world setting than the common closed-set classification that assumes all labeled and unlabeled data belong to the same classes.

\begin{figure*}[h]
  \begin{center}
\includegraphics[width=\textwidth]{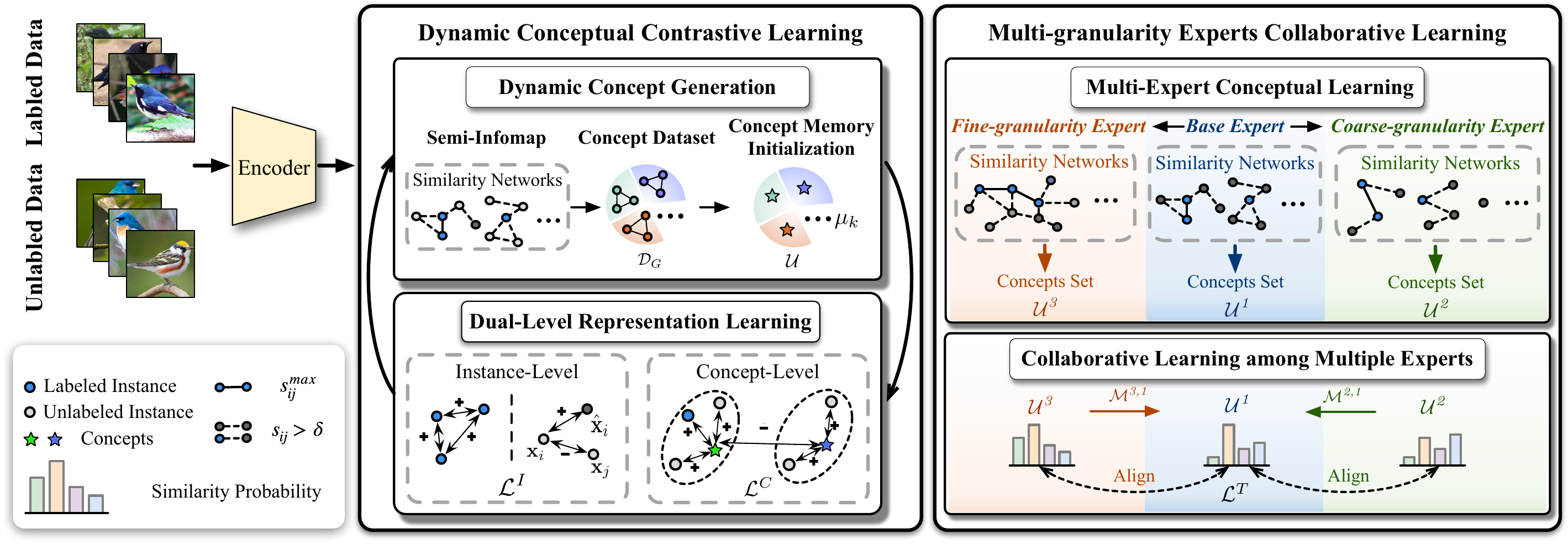}
\caption{
The MGCE framework consists of two principal modules: Dynamic Conceptual Contrastive Learning (DCCL) and Multi-Granularity Experts Collaborative Learning (MECL). DCCL alternates between Dynamic Concept Generation (DCG) and Dual-level Representation Learning DRL) to form a conceptual expert. Building on DCCL, MECL integrates two components: Multi-Expert Conceptual Learning (ME), which constructs expert branches with varying levels of information granularity, and Collaborative Learning (CL), which aligns and integrates their insights. This enables experts to be optimized collectively and to mutually benefit from each another. 
}
\label{fig:framework}
  \end{center}
\end{figure*}

\noindent \textbf{Framework Overview.} 
To tackle the GCD problem, we propose a novel framework, \textit{MGCE}, illustrated in Fig.~\ref{fig:framework}. The framework consists of two modules: Dynamic Conceptual Contrastive Learning (DCCL) and Multi-Granularity Experts Collaborative Learning (ME-CL). The DCCL module alternates between two components: Dynamic Concept Generation (DCG) and Dual-level Representation Learning (DRL). DCG leverages a Semi-Infomap algorithm to assign concept labels to the entire training dataset $\mathcal{D}$, while DRL performs instance- and concept-level representation learning based on these labels to learn more discriminative features. This iterative alternation builds a category discovery expert, progressively enhancing the model’s ability to uncover novel categories. The MECL module consists of two components: Multi-Expert Conceptual Learning (ME) and Collaborative Learning among Multiple Experts (CL). Building on the discovery expert developed in DCCL, the ME component introduces additional experts with finer and coarser information granularities, while the CL component further facilitates their collaboration. Together, these mechanisms promote alignment in category understanding across different granularities.

\subsection{Dynamic Conceptual Contrastive Learning}
The goal of GCD is to cluster unlabeled data by leveraging categorical knowledge acquired from labeled data. Prior GCD studies~\cite{gcd,simgcd,dccl} highlight that learning discriminative representations is essential for achieving accurate clustering. To this end, we propose a comprehensive discriminative learning framework, DCCL, which jointly incorporates instance-level and concept-level representation learning.

\subsubsection{Instance-Level Representation Learning}
\label{sec:instance_level}

\par
Following prior GCD methods~\cite{gcd,dccl,simgcd}, we adopt the self-supervised pre-trained model DINO~\cite{dino} as the backbone representation extractor. We then fine-tune the backbone on both labeled and unlabeled data to adapt it to the target dataset and enhance its representations.

\noindent \textbf{Instance-Level Contrastive Objective.} 
Following prior work~\cite{gcd}, we formulate the instance-level contrastive objective by combining a self-supervised contrastive loss over all samples with a supervised contrastive loss over labeled samples. Let $\mathcal{E}(\cdot)$ denote the image encoder and $\phi(\cdot)$ the MLP projection head. For an input image $\mathbf{x}_i$, the projected representation is $\mathbf{h}_i = \phi(\mathcal{E}(\mathbf{x}_i))$. Assume $\mathbf{x}_i$ and $\hat{\mathbf{x}}_i$ represent two augmented views of the same image within a randomly sampled mini-batch $\mathcal{B}$. 
The instance-level contrastive loss $\mathcal{L}_{con}$ is defined as:

\begin{equation}
\begin{aligned}
&\mathcal{L}_{con} = -(1-\lambda)\,\frac{1}{|\mathcal{B}|} 
   \sum\limits_{i \in \mathcal{B}}
   \log \frac{\exp(\langle h_i, \hat{h}_i \rangle / \tau_u)}
   {\sum\limits_{j \in \mathcal{B} \setminus \{i\}} \exp(\langle h_i, h_j \rangle / \tau_u)} \\
& - \lambda\,\frac{1}{|\mathcal{B}_L|}
   \sum\limits_{i \in \mathcal{B}_L}
   \frac{1}{|\mathcal{P}(i)|} 
   \sum\limits_{q \in \mathcal{P}(i)}
   \log \frac{\exp(\langle h_i, h_q \rangle / \tau_l)}
   {\sum\limits_{j \in \mathcal{B}_L \setminus \{i\}} \exp(\langle h_i, h_j \rangle / \tau_l)}.
\label{loss_con}
\end{aligned}
\end{equation}
where $\mathcal{B}_L$ is the labeled subset of $\mathcal{B}$, and $\mathcal{P}(i)$ denotes the set of indices of samples that share the same class label as anchor $\mathbf{x}_i$. 
The parameters $\tau_u$ and $\tau_l$ are the temperature values, and $\lambda$ is a balancing factor that adjusts the contributions of self-supervised contrastive learning (the former term in Eq.~\ref{loss_con}) and supervised contrastive learning (the latter term in Eq.~\ref{loss_con}).

\noindent \textbf{Classification Loss with Switchable Classifier.} 
To effectively exploit the categorical knowledge from labeled data, we introduce a supervised classification loss:
\begin{equation}
\mathcal{L}_{cls}^s = \frac{1}{|\mathcal{B}_L|} \sum_{i \in \mathcal{B}_L} \ell(y_i, \mathbf{p}_i),
\label{loss_cls_sup}
\end{equation}
where $\ell$ is the cross-entropy loss and $\mathbf{p}_i = \sigma(f_{z}(\mathbf{z}_i)/\tau_s)$ denotes the predicted probability distribution of $\mathbf{x}_i$. Here, $f_{z}$ is a parametric classifier applied only to labeled samples, with output dimension $K_L$, $\sigma(\cdot)$ is the softmax function, and $\tau_s$ is a temperature parameter. 
When the number of categories in the unlabeled set ($K_U$) is unknown, the classification loss is computed solely on labeled data, yielding 
$\mathcal{L}_{cls} = \lambda \mathcal{L}_{cls}^s$, 
where $\lambda$ is a balancing factor consistent with that in $\mathcal{L}_{con}$.

\par
Furthermore, \textbf{\textit{when the number of classes in the unlabeled dataset is known,}} 
we adopt the unsupervised objective introduced in SimGCD~\cite{simgcd} for consistency in comparison:
\begin{equation}
\mathcal{L}_{cls}^u = \frac{1}{|\mathcal{B}|} \sum_{i \in \mathcal{B}} \ell(\hat{\mathbf{q}}_i, \mathbf{p}_i) - \epsilon H(\hat{\mathbf{p}}),
\label{loss_cls_unsup}
\end{equation}
where $\hat{\mathbf{q}}_i$ denotes the sharpened prediction for $\hat{\mathbf{x}}_i$ with temperature $\tau_t$. 
$H(\hat{\mathbf{p}}) = -\sum_k \hat{\mathbf{p}}^{(k)} \log \hat{\mathbf{p}}^{(k)}$ is introduced to avoid trivial solutions, and $\hat{\mathbf{p}} = \tfrac{1}{2|\mathcal{B}|} \sum_{i \in \mathcal{B}} (\mathbf{p}_i + \hat{\mathbf{p}}_i)$ 
denotes the mean prediction across two augmented views $\mathbf{x}_i$ and $\hat{\mathbf{x}}_i$.
The hyperparameter $\epsilon$ is set exactly as in SimGCD~\cite{simgcd}.
The final classification loss is defined as 
$\mathcal{L}_{cls} = (1-\lambda)\mathcal{L}_{cls}^u + \lambda\mathcal{L}_{cls}^s$.

\par
In summary, to accommodate different scenarios, we define the final classification loss function $\mathcal{L}_{cls}$ as follows:
\begin{equation}
\mathcal{L}_{cls} =
\begin{cases}
\lambda\, \mathcal{L}_{cls}^s, & \text{if } K_U \text{ is unknown}, \\[1ex]
(1-\lambda)\,\mathcal{L}_{cls}^u + \lambda\, \mathcal{L}_{cls}^s, & \text{if } K_U \text{ is known}.
\end{cases}
\end{equation}

\noindent \textbf{Instance-Level Total Loss.} The instance-level representation learning loss, $\mathcal{L}^I$, is defined as the sum of the contrastive and classification losses:
\begin{equation}
\mathcal{L}^I = \mathcal{L}_{con} + \mathcal{L}_{cls}.
\label{loss_instance}
\end{equation}

\noindent \textbf{Discussion.} 
While instance-level contrastive learning enhances the clustering capability of GCD models, it fails to capture the intrinsic relationships among instances belonging to the same concept (e.g., classes, superclasses, and subclasses), which may result in suboptimal feature representations. 
To overcome this limitation, we propose an alternating contrastive learning framework that incorporates dynamic concept generation and concept-level representation learning. 
The proposed framework leverages concept-level knowledge embedded in unlabeled data to improve the discriminability of learned representations.

\subsubsection{Dynamic Concept Generation}
As illustrated in Fig.~\ref{fig:highlight}~(a), the discriminative capability of the model improves progressively during training. To ensure consistency between dynamic model optimization and concept knowledge mining, we propose a Dynamic Concept Generation (DCG) module, which dynamically assigns concept labels to both labeled and unlabeled data at each training epoch based on the model’s representation capacity. Specifically, DCG employs a Semi-Infomap clustering algorithm to generate concept labels and maintains a concept level memory buffer for subsequent concept-level representation learning.

\noindent \textbf{Semi-Infomap clustering algorithm.} Vaze et al.~\cite{gcd} introduced a semi-supervised $k$-means algorithm for evaluation under a fixed number of categories. 
Since the actual number of clusters is often unknown during training, we adopt the Infomap algorithm~\cite{infomap}, which requires no prior knowledge of the number of clusters, to dynamically generate concepts. 
For an input image $\mathbf{x}_i$, its feature representation is denoted as $\mathbf{z}_i = \mathcal{E}(\mathbf{x}_i)$. 
We then construct a feature similarity network $\{\mathbf{z}_i\}_{i=1}^{M+N} \in \mathcal{Z}$, where edges are weighted by pairwise feature similarities. 
Within this network, Infomap partitions semantically coherent communities based on structural connectivity, thereby identifying meaningful concepts in $\mathcal{Z}$. 
To further incorporate supervision from labeled data, a novel Semi-Infomap clustering algorithm is developed, which integrates similarity constraints derived from labeled samples into the network. 
Formally, the similarity $s_{ij}$ between the $i$-th and $j$-th features is defined as:
\begin{equation}\label{eq:s_ij}
    s_{ij}=[(\mathbf{z}_{i}/{\left \| \mathbf{z}_{i} \right \| })\bm{\cdot} (\mathbf{z}_{j}/\left \| \mathbf{z}_{j} \right \|) + 1]/2\in [0,1].
\end{equation}

Next, we construct an adjacency matrix $\mathcal{A}$ to encode the connectivity among all samples. 
The edge weight between the $i$-th and $j$-th samples is defined as:
\begin{equation}\label{eq:adjacent}
    \mathcal{A}_{ij} = \begin{cases}
    s^{max}_{i}, & \text{if } (y_{i} = y_{j}) \wedge (y_{i}, y_{j} \in \mathcal{Y}_{L})\\ 
    s_{ij}, & \text{if } (y_{i} \in \mathcal{Y}_{U} \vee y_{j} \in \mathcal{Y}_{U}) \wedge (s_{ij} > \delta)\\
    0, & \text{otherwise}
    \end{cases},
\end{equation}
\begin{equation}\label{eq:max_s}
    s^{\text{max}}_i = \max \left\{ s_{ij} \mid j \in \{1, \dots, M+N\} \right\},
\end{equation}
where $\delta$ is a threshold used to retain only high-confidence connections. 
For positive pairs, the similarity is assigned the maximum value among their neighborhood similarities. To more effectively capture semantic similarity between instances and identify semantically coherent concepts, 
we restrict the adjacency matrix $\mathcal{A}'$ to high-quality local neighborhoods:
\begin{equation}
\label{eq:adjacency}
\mathcal{A}'_{ij} =
\begin{cases}
\mathcal{A}_{ij}, & \text{if } j \in \operatorname{KNN}(i, k_{\text{nn}}), \\
0, & \text{otherwise},
\end{cases}
\end{equation}
where $\operatorname{KNN}(i, k_{\text{nn}})$ denotes the set of $k_{\text{nn}}$ nearest neighbors of instance $i$. 
The parameter $k_{\text{nn}}$ is influenced by both the scale of the similarity network and the density of edge connections. 
To select an appropriate value, we adopt an adaptive strategy guided by the labeled subset. 
Specifically, we construct the similarity network jointly using labeled and unlabeled samples, and then evaluate candidate values of $k_{\text{nn}}$ 
based on clustering accuracy on labeled data and the quality of the estimated concept number. The optimal $k_{\text{nn}}$ is determined via a coarse-to-fine search. 
In the coarse stage, candidate values are sampled at logarithmic intervals to efficiently cover the search space. 
In the fine stage, both clustering accuracy and the number of estimated concepts are used as criteria to select the best $k_{\text{nn}}$. 
The complete procedure is outlined in Algorithm~\ref{alg:knn_selection}.

Once the $k_{\text{nn}}$ is adaptively determined and the adjacency matrix $\mathcal{A}'$ is constructed, Infomap is employed to partition the similarity network $\mathcal{Z}$ into structurally coherent communities, which are treated as the discovered concepts.

\begin{algorithm}[t]
\caption{Adaptive $k_{\text{nn}}$ Selection}
\label{alg:knn_selection}
\KwIn{%
Labeled data $\mathcal{D}_L$, 
Unlabeled data $\mathcal{D}_U$, 
}
\KwOut{Optimal neighbor number $k^*_{\text{nn}}$}

\BlankLine
\textbf{/*Coarse Search*/}\\
Coarse candidate set $k^c_{\text{nn}} = \{2^n \mid n=2,\dots,9\},$ \\
\For{$k_{\text{nn}} \in k^c_{\text{nn}}$}{
    Perform Semi-Infomap with top-$k_{\text{nn}}$ neighbors\;
    Evaluate clustering accuracy (ACC) on $\mathcal{D}_L$\;
}
Select best coarse $k^c_{\text{nn}}{}^* = \arg\max \text{ACC}(k_{\text{nn}})$\;

\BlankLine
\textbf{/*Fine Search*/}\\
Define a candidate set $k^f_{\text{nn}}$ as the integer interval 
between the neighboring coarse values around $k^c_{\text{nn}}{}^*$ (with step size 10 or 50 if large)\;
\For{$k_{\text{nn}} \in k^f_{\text{nn}}$}{
    Perform Semi-Infomap with top-$k_{\text{nn}}$ neighbors\;
    Compute clustering accuracy $\text{ACC}(k_{\text{nn}})$\ on $\mathcal{D}_L$\;
    Compute estimated cluster number $K_{\text{est}}$ on $\mathcal{D}_L$\;
    Calculate error rate 
    $\text{ErrRate} = \frac{|K_{\text{true}} - K_{\text{est}}|}{K_{\text{est}}}$\;
    Compute error-adjusted accuracy 
    $\text{ERR\_ACC}(k_{\text{nn}}) = \text{ACC}(k_{\text{nn}}) \times (1-\text{ErrRate})$\;
}
Select $k^*_{\text{nn}} = \arg\max_{k_{\text{nn}} \in k^f_{\text{nn}}} \text{ERR\_ACC}(k_{\text{nn}})$\;

\Return $k^*_{\text{nn}}$\;
\end{algorithm}

\noindent \textbf{Concept Memory Initialization.} 
Samples within the same sub-network are assumed to share a common concept. 
Accordingly, we define the generated concept label set for both labeled and unlabeled instances as $\mathcal{C} = \{c_{i} \in \mathcal{Y}_{G} \}^{N+M}_{i=1}$, where $\mathcal{Y}_G$ denotes the generated concept label space. The extracted feature vectors are then paired with corresponding concept labels to construct 
the generated concept dataset $\mathcal{D}_{G} = \{(\mathbf{z}_{i}, c_{i})\}^{N+M}_{i=1}  \in \mathcal{Z} \times \mathcal{Y}_{G}$. A concept-level memory buffer is maintained to store dynamic concept representations for contrastive learning. 
The initialization involves computing the mean feature vector of samples assigned to the same concept label, thereby forming distinct concept representations. 
Formally, the initial concept representation set $\mathcal{U}$ is defined as:
\begin{equation}\label{eq:initialization}
\mathcal{U} = \{\bm{\mu}_{k}\}_{k=1}^{K_G}, \quad 
\bm{\mu}_{k} = \tfrac{1}{|\mathcal{D}_{G}^{k}|} \sum_{\mathbf{z}_{i} \in \mathcal{D}_{G}^{k}} \mathbf{z}_{i}, \quad 
K_G = |\mathcal{Y}_{G}|,
\end{equation}
where $\mathcal{D}_{G}^{k}$ denotes the $k$-th concept subset, meaning that $\mathbf{z}_{i} \in \mathcal{D}_{G}^{k}$ indicates $c_i = k$. 
The concept memory buffer is re-initialized at every epoch using simple center-cropped images for stability, while the number of concepts $K_G$ is dynamically updated throughout training.

\subsubsection{Concept-Level Representation Learning}
\label{sec:concept_level}

\noindent\textbf{Concept-Level Contrastive Learning.}
Leveraging the generated concept representations, we introduce a concept-level contrastive learning objective. Specifically, we sample $N_{C}$ concept labels and select a fixed number $N_{I}$ of instances for each label, forming a mini-batch $\mathcal{B}_{C}$ comprising $N_{C} \times N_{I}$ samples. Each projected instance representation $\mathbf{v}_i = \phi_C(\mathbf{z}_i)$ is then contrasted against all concept prototypes. The objective is to align each instance with its corresponding concept prototype while discouraging similarity with other prototypes. Formally, the concept-level contrastive loss is defined as:
\begin{equation}
\mathcal{L}^{C} = \frac{1}{|\mathcal{B}_C|} \sum_{i \in \mathcal{B}_C} -\log \frac{\exp \left(\langle\mathbf{v}_i \cdot \bm{\mu}_{c_i}\rangle / \tau_{c}\right)}{\sum_{k=1}^{K_G} \exp(\langle\mathbf{v}_i \cdot \bm{\mu}_k\rangle / \tau_{c})}
\end{equation}
where $\tau_{c}$ is a temperature hyper-parameter.

\noindent\textbf{Concept Memory Update.}
In contrast to prior approaches~\cite{he2020momentum,pcl}, which maintain memory banks of all training instances, our method stores only the concept representations, thereby substantially reducing storage requirements. Moreover, updating memory in an instan-ce-wise manner may induce inconsistencies across training iterations. To address this issue, we re-sample $N_C$ concept labels and select $N_I$ instances per label to form the mini-batch $\mathcal{B}_C$, as described above. The concept representation corresponding to each instance is then updated using a momentum-based strategy:
\begin{equation}\label{eq:update}
\bm{\mu}_{{c}_{i}} \leftarrow \eta \bm{\mu}_{\mathbf{c}_{i}}+(1-\eta)\mathbf{v}_{i},
\end{equation}
where $\eta$ is the momentum updating factor. 

\noindent\textbf{Discussion.} The proposed DCCL enhances representation learning by alternating between dynamic concept generation and dual-level contrastive learning, thereby enabling accurate category discovery. However, DCCL generates conceptual knowledge at a single granularity across training epochs, but lacks the capacity to integrate multiple granularities within the same epoch. This limitation results in an incomplete conceptual understanding. For instance, at a coarser superclass level, features of categories belonging to the same superclass are expected to be more similar, whereas at a finer subclass level, a single category may encompass diverse morphological variations. To overcome this limitation, we generalize the single-expert DCCL into a multi-expert learning paradigm that simultaneously generates and integrates concepts across multiple granularities within each epoch, thereby facilitating a more holistic and robust conceptual representation.

\subsection{Multi-Granularity Experts Collaborative Learning}
\noindent\textbf{Insights.} 
When generating concept labels with the Semi-Infomap algorithm, we focus on high-quality neighboring samples as determined by the parameter $k_{\text{nn}}$. 
This parameter serves as a search radius that governs both the connectivity among samples and the granularity of information captured during concept generation. 
A smaller $k_{\text{nn}}$ restricts connections to the most similar samples, yielding more sub-networks and finer-grained concepts. 
Conversely, a larger $k_{\text{nn}}$ includes broader neighborhoods, producing fewer sub-networks and coarser-grained concepts. 
These granularities are interdependent and complementary: fine granularity provides precise distinctions among closely related classes, whereas coarse granularity captures higher-level semantic relationships. 
Integrating these perspectives enables the mo-del to construct hierarchical class structures, thereby enhancing semantic understanding, especially when discriminating among visually similar categories. 
Motivated by this insight, we introduce additional experts at both coarse and fine granularities by strategically adjusting $k_{\text{nn}}$, thus equipping the model with hierarchical semantic perspectives.

\subsubsection{Multi-Expert Conceptual Learning}
\noindent\textbf{Multiple Concept Experts.} 
By varying the parameter $k_{\text{nn}}$, concept experts can be constructed to operate at different levels of granularity. 
A base expert, denoted as \textit{Expert 1}, is defined with a nearest-neighbor parameter $k_{\text{nn}}^1$, 
where $k_{\text{nn}}^1$ corresponds to the adaptively selected value (Algorithm~\ref{alg:knn_selection}). 
Using this reference expert as the basis, additional experts are derived through a scaling factor $\mathcal{R} \in (0,1)$. 
Specifically, we define \textit{Expert 2} and \textit{Expert 3} with nearest-neighbor parameters:
\begin{equation}
k_{\text{nn}}^2 = k_{\text{nn}}^1 \cdot \mathcal{R}, \quad 
k_{\text{nn}}^3 = k_{\text{nn}}^1 / \mathcal{R}.
\label{eq:scaling}
\end{equation}
\textit{Expert 2}, operating on fewer neighbors, captures finer-grained semantic distinctions, 
whereas \textit{Expert 3}, operating on more neighbors, emphasizes coarser-grained structures. 

\noindent\textbf{Multi-Expert Contrastive Learning.} 
For each expert, the concept label set generated by the Semi-Infomap algorithm is denoted as 
$\mathcal{C}^r = \{c_{i}^r\}_{i=1}^{N+M} \in \mathcal{Y}_{G}^r$, 
and the corresponding concept representation set is 
$\mathcal{U}^r = \{\bm{\mu}_{k}^r\}_{k=1}^{K_G^r}$, 
where $r \in \{1,2,3\}$. 
The projected feature for each instance is obtained as 
$\mathbf{v}_i^r = \phi_C^r(\mathbf{z}_i)$. 
Based on these definitions, the concept-level contrastive loss for expert $r$ is formulated as:
\begin{equation}
\mathcal{L}^{C,r} 
= \frac{1}{|\mathcal{B}_C|} 
\sum_{i \in \mathcal{B}_C} 
-\log 
\frac{\exp \left(\langle \mathbf{v}_i^r, \bm{\mu}_{c_i}^r \rangle / \tau_{c}\right)}
{\sum_{k=1}^{K_G^r} \exp \left(\langle \mathbf{v}_i^r, \bm{\mu}_{k}^r \rangle / \tau_{c}\right)}.
\end{equation}
The overall concept-level contrastive loss is then defined as the aggregation over all experts:
\begin{equation}
\mathcal{L}^C = \sum_{r=1}^{3} \mathcal{L}^{C,r}.
\end{equation}

\noindent\textbf{Discussion.} 
Multi-expert concept learning, which integrates insights from both fine- and coarse-grained categories, provides complementary perspectives and facilitates a more comprehensive semantic understanding. 
However, the independent operation of these experts prevents effective interaction across different granularity levels. 
This isolation results in a lack of shared knowledge that could otherwise strengthen category understanding, thereby constraining mutual knowledge exchange and collaborative improvement.

\subsubsection{Collaborative Learning among Multiple Experts}
\noindent Interactions across different levels of granularity can be mutually beneficial. 
Coarse-grained understanding captures shared characteristics among categories, thereby facilitating the identification of high-level similarities between known and novel classes. 
In contrast, fine-grained understanding captures subtle intra-class morphological variations, which are crucial for delineating precise category boundaries. 
Integrating these perspectives facilitates a unified understanding of novel categories, thereby ensuring stable and consistent category comprehension across different levels. 
In our framework, the concept representation set produced by each expert serves as prototypes for categories at its respective granularity. 
By computing pairwise similarities between these prototype sets, knowledge can be effectively transferred across experts, promoting interaction and alignment among different granularities.

\noindent\textbf{Concept Alignment Matrix.} 
Since \textit{Expert 2} and \textit{Expert 3} are derived from \textit{Expert 1}, we designate \textit{Expert 1} as the reference expert and analyze its interactions with the other two experts throughout training. 
To this end, we introduce a concept alignment matrix to quantify the relationships between the concept representations of different experts:
\vspace{0.3cm}
\begin{equation}
\mathcal{M}^{r,1} \in \mathbb{R}^{K_G^r \times K_G^1}, \quad 
\mathcal{M}^{r,1}_{ij} = \frac{\bm{\mu}_i^r \cdot \bm{\mu}_j^1}{\|\bm{\mu}_i^r\| \|\bm{\mu}_j^1\|}, \quad r \in \{2,3\}.
\end{equation}
Each entry $\mathcal{M}^{r,1}_{ij}$ corresponds to the cosine similarity between the $i$-th concept representation of expert $r$ and the $j$-th concept of the reference expert, thereby capturing semantic correspondences across different granularities.

\noindent\textbf{Multi-Expert Collaborative Learning.} 
For each expert, given the projected representation $\mathbf{v}_i^r$ and the concept representation set $\mathcal{U}^r = \{\bm{\mu}_{k}^r\}$, 
the similarity distribution between the $i$-th sample and the concepts is defined as:
\vspace{0.3cm}
\begin{equation}
\hat{\mathbf{p}}_i^r = \sigma\!\left( \big[ \cos(\mathbf{v}_i^r, \bm{\mu}_1^r), \ldots, \cos(\mathbf{v}_i^r, \bm{\mu}_{K_G^r}^r) \big] \right),
\end{equation}
where $\sigma(\cdot)$ denotes the softmax function and $\cos(\cdot)$ the cosine similarity. 
To enable cross-granularity interaction, \textit{Expert 2} and \textit{Expert 3} project their similarity distributions onto the concept space of \textit{Expert 1} using the alignment matrix:
\vspace{0.3cm}
\begin{equation}
\hat{\mathbf{p}}_i^{r \rightarrow 1} = \hat{\mathbf{p}}_i^r \mathcal{M}^{r,1}, \quad r \in \{2,3\}.
\end{equation}
We then enforce consistency between $\hat{\mathbf{p}}_i^1$ and $\hat{\mathbf{p}}_i^{r \rightarrow 1}$ through a symmetric KL divergence, yielding the collaborative loss:
\vspace{0.3cm}
\begin{equation}
\mathcal{L}^T = \sum_{r=2}^{3} \tfrac{1}{2} \Big( 
D_{KL}(\hat{\mathbf{p}}_i^1 \parallel \hat{\mathbf{p}}_i^{r \rightarrow 1}) + 
D_{KL}(\hat{\mathbf{p}}_i^{r \rightarrow 1} \parallel \hat{\mathbf{p}}_i^1) 
\Big),
\end{equation}
where $D_{KL}(\cdot \parallel \cdot)$ denotes the Kullback–Leibler divergence. 
This bidirectional consistency promotes knowledge transfer across experts of different granularities.

\subsection{Training and Inference}
\noindent\textbf{Training.} 
During the training process, the concept-level contrastive loss and the collaborative learning loss are treated as a unified concept-level component. The total loss is formulated as:
\begin{equation}\label{eq:total_loss}
    \mathcal{L}_{\text{total}} = \mathcal{L}^{I} + \alpha (\mathcal{L}^{C} + \mathcal{L}^{T}),
\end{equation}
where $\alpha$ is a weighting factor that balances the contributions of the instance-level and concept-level learning objectives. The pseudo-code of \textit{MGCE} is provided in Algorithm~\ref{algorithm:DCCL}.

\begin{algorithm}[t]
\caption{Pseudocode of MGCE}
\label{algorithm:DCCL}
\SetAlgoLined
\KwIn{Labeled data $\mathcal{D}_{L}$, unlabeled data $\mathcal{D}_{U}$, feature extractor $\mathcal{E}(\cdot)$, projection heads $\phi(\cdot)$ and $\phi_C^r(\cdot)$ $(r=1,2,3)$, and classifier $f_z$.}
\KwOut{Updated $\mathcal{E}(\cdot)$, $\phi(\cdot)$, $\phi_C^r(\cdot)$.}

\textbf{/*Global initialization*/} \\
Obtain $k_{\text{nn}}$ via Algorithm~\ref{alg:knn_selection}\;
Assign expert-specific parameters $\{k_{\text{nn}}^r\}_{r=1}^3$ by Eq.~\ref{eq:scaling}\;

\BlankLine
\For{$n = 1$ \textbf{to} $max\_epoch$}{
  \textbf{/*Concept generation and memory initialization*/} \\
  \For{each expert $r$}{
      Compute projected features $\mathbf{v}_i^r = \phi_C^r(\mathbf{z}_i)$\;
      Construct adjacency matrix $\mathcal{A}^{\prime r}$ by Eq.~\ref{eq:adjacency}\;
      Apply Semi-Infomap clustering to obtain concept labels $\mathcal{C}^r$\;
      Initialize concept memory buffer using Eq.~\ref{eq:initialization}\;
  }

  \BlankLine
  \textbf{/*Optimization*/} \\
  \For{$t = 1$ \textbf{to} $max\_iteration$}{
      Sample mini-batches from $\mathcal{D}_{L} \cup \mathcal{D}_{U}$\;
      Compute total loss $\mathcal{L}_{\text{total}}$ by Eq.~\ref{eq:total_loss}\;
      Update $\mathcal{E}(\cdot)$, $f_z$, $\phi(\cdot)$, and $\phi_C^r(\cdot)$ via SGD\;
      Update concept memory buffer using Eq.~\ref{eq:update}\;
  }
}
\end{algorithm}

\noindent\textbf{Inference.} 
At test time, when the number of categories is unknown, we apply Semi-Infomap clustering with the adaptive parameter $k_{\text{nn}}$ obtained via Algorithm~\ref{alg:knn_selection} to partition the unlabeled dataset $\mathcal{D}_U$. 
When the number of categories is known but the discovered number of clusters exceeds the ground-truth, we iteratively merge the smallest clusters into their most similar counterparts until convergence. 
The final partition is then used as the category assignment result.

\noindent\textbf{Estimating the Number of Classes.} 
Unlike non-parametric GCD methods~\cite{gcd,promptcal,gpc,pim,cms}, our MGCE framework integrates category discovery into the training procedure, thereby embedding category number estimation implicitly within the inference process. 
As a result, the estimated number of categories is directly obtained from the final partitioning outcome. 
When the number of categories is unknown, we impose a minimum cluster-size constraint to improve robustness: clusters with fewer than four samples are excluded from being considered valid categories, as such small clusters are typically attributed to noise or insignificant patterns rather than meaningful semantics.

\section{Experiments}

\subsection{Experimental Setup}
\par\textbf{Datasets.} We evaluate on nine fine-grained benchmarks: CUB~\cite{cub}, Stanford Cars~\cite{scars}, NABirds~\cite{nabirds}, Herbarium19~\cite{herb19}, and five iNaturalist superclasses—Fungi, Animalia, Mollusca, Actinopterygii and Reptilia~\cite{inaturalist}.
We also report results on the generic datasets CIFAR-100~\cite{cifar} and ImageNet-100~\cite{imagenet}.
Following~\cite{gcd}, a subset of classes is selected as the labeled set $\mathcal{Y}_L$.
From these classes, 50\% of the images form the labeled split $\mathcal{D}_L$, while the remaining images, together with all images from the unlabeled classes, constitute the unlabeled split $\mathcal{D}_U$. Tab.~\ref{tab:dataset} summarizes dataset statistics.

\begin{table}[h]
\centering
\caption{Dataset statistics and splits.\label{tab:dataset}}
\setlength{\tabcolsep}{1pt}
\renewcommand{\arraystretch}{0.95}
\begin{tabular}{lrrrr}
\toprule
\multirow{2}{*}{Dataset} & \multicolumn{2}{c}{Labeled} & \multicolumn{2}{c}{Unlabeled} \\
\cmidrule(lr){2-3}\cmidrule(l){4-5}
& \#~Images & \#~Classes & \#~Images & \#~Classes \\
\midrule
\multicolumn{5}{l}{\textit{Small\mbox{-}scale datasets}} \\
\midrule
CUB~\cite{cub}                     &  1,498 & 100  &  4,496 & 200 \\
Stanford Cars~\cite{scars}         &  2,000 &  98  &  6,144 & 196 \\
Animalia~\cite{inaturalist}        &  1,492 &  39  &  5,098 &  77 \\
Fungi~\cite{inaturalist}           &  1,786 &  61  &  5,820 & 121 \\
Mollusca~\cite{inaturalist}        &  2,417 &  47  &  6,960 &  93 \\
Actinopterygii~\cite{inaturalist}  &    576 &  27  &  2,043 &  53 \\
\midrule
\multicolumn{5}{l}{\textit{Large\mbox{-}scale datasets}} \\
\midrule
NABirds~\cite{nabirds}             &  5,627 & 278  & 18,302 & 555 \\
Herbarium19~\cite{herb19}          &  8,869 & 341  & 25,356 & 683 \\
Reptilia~\cite{inaturalist}         & 9,143 & 145 & 31,738& 289 \\
CIFAR\mbox{-}100~\cite{cifar}      & 20,000 &  80  & 30,000 & 100 \\
ImageNet\mbox{-}100~\cite{imagenet}& 31,860 &  50  & 95,255 & 100 \\
\bottomrule
\end{tabular}
\end{table}

\begin{table*}[th]
\small
\centering
\caption{Comparison on small-scale fine-grained datasets. The best results are highlighted in \textbf{bold}, and `--' indicates results not reported due to unavailable code.}

\label{tab:sota_1}
\setlength{\tabcolsep}{1pt}
\renewcommand{\arraystretch}{0.6}
\begin{tabular}{@{}l|ccc|ccc|ccc|ccc|ccc|ccc|ccc@{}}
\toprule
\multirow{2}{*}{Method}
& \multicolumn{3}{c|}{CUB}
& \multicolumn{3}{c|}{Stanford Cars}
& \multicolumn{3}{c|}{Animalia}
& \multicolumn{3}{c|}{Fungi}
& \multicolumn{3}{c|}{Mollusca}
& \multicolumn{3}{c|}{Actinopterygii}
& \multicolumn{3}{c}{Average} \\
\cmidrule(lr){2-4}
\cmidrule(lr){5-7}
\cmidrule(lr){8-10}
\cmidrule(lr){11-13}
\cmidrule(lr){14-16}
\cmidrule(lr){17-19}
\cmidrule(l){20-22}
& All & Old & New
& All & Old & New
& All & Old & New
& All & Old & New
& All & Old & New
& All & Old & New
& All & Old & New \\
\midrule
\multicolumn{22}{l}{(a) \textit{Comparison when the number of classes $K_U$ in the unlabeled dataset is known}} \\
\midrule
GCD~\cite{gcd}
& 51.3 & 56.6 & 48.7
& 39.0 & 57.6 & 29.9
& 48.3 & 69.0 & 39.8
& 48.7 & 68.3 & 40.0
& 51.2 & 67.0 & 42.9
& 39.9 & 52.1 & 35.1
& 46.4 & 61.8 & 39.4 \\

XCon~\cite{xcon}
& 52.1 & 54.3 & 51.0
& 40.5 & 58.8 & 31.7
& 51.9 & 66.7 & 45.8
& 50.3 & 65.4 & 43.6
& 52.1 & 70.2 & 42.4
& 42.8 & 50.9 & 39.6
& 48.3 & 61.1 & 42.4 \\

GPC~\cite{gpc}
& 55.4 & 58.2 & 53.1
& 42.8 & 59.2 & 32.8
& 46.8 & 70.9 & 36.8
& 42.7 & 63.3 & 33.6
& 45.9 & 64.1 & 36.3
& 42.0 & 47.9 & 38.3
& 45.9 & 60.6 & 38.5 \\

InfoSieve~\cite{infosieve}
& 69.4 & \textbf{77.9} & 65.2
& 55.7 & 74.8 & 46.4
& 50.7 & 71.1 & 42.2
& 47.4 & 63.0 & 40.5
& 48.0 & 66.7 & 38.0
& 41.9 & 52.6 & 37.7
& 52.2 & 67.7 & 45.0 \\

SimGCD~\cite{simgcd}
& 60.3 & 65.6 & 57.7
& 53.8 & 71.9 & 45.0
& 45.1 & 74.4 & 32.9
& 43.3 & 56.4 & 37.5
& 48.1 & 70.6 & 36.1
& 42.1 & 52.1 & 38.2
& 48.8 & 65.2 & 41.2 \\

PIM~\cite{pim}
& 62.7 & 75.7 & 56.2
& 43.1 & 66.9 & 31.6
& 45.6 & 76.2 & 33.0
& 47.1 & 70.8 & 36.6
& 47.0 & 68.8 & 35.4
& 41.8 & 52.1 & 37.7
& 47.9 & 68.4 & 38.4 \\

CMS~\cite{cms}
& 68.2 & 76.5 & 64.0
& 56.9 & 76.1 & 47.6
& 54.0 & \textbf{77.5} & 44.3
& 51.3 & \textbf{71.3} & 42.4
& 52.1 & \textbf{74.0} & 40.5
& \textbf{47.0} & \textbf{62.5} & 41.0
& 54.9 & \textbf{73.0} & 46.6 \\

SPTNet~\cite{sptnet}
& 65.8 & 68.8 & 65.1
& 59.0 & \textbf{79.2} & 49.3
& 47.5 & 60.2 & 42.2
& 44.2 & 63.0 & 35.8
& 47.9 & 61.4 & 40.7
& 41.9 & 53.1 & 37.4
& 51.0 & 64.3 & 45.1 \\

LegoGCD~\cite{legogcd}
& 63.8 & 71.9 & 59.8
& 57.3 & 75.7 & 48.4
& 45.9 & 64.1 & 38.4
& 42.7 & 62.8 & 33.8
& 49.4 & 65.3 & 40.9
& 42.6 & 51.4 & 39.2
& 50.3 & 65.2 & 43.4 \\

ProtoGCD~\cite{protogcd}
& 63.2 & 68.5 & 60.5
& 53.8 & 73.7 & 44.2
& 49.3 & 67.8 & 41.6
& 45.2 & 66.2 & 35.9
& 50.3 & 70.6 & 39.5
& 43.9 & 55.9 & 39.1
& 50.9 & 67.1 & 43.5 \\

ConGCD~\cite{congcd}
& 61.6 & 65.1 & 59.5
& 54.5 & 72.2 & 47.8
& -- & -- & --
& -- & -- & --
& -- & -- & --
& -- & -- & --
& -- & -- & -- \\

APL~\cite{aplgcd}
& 64.5 & 68.1 & 62.1
& 60.1 & 77.6 & 51.2
& -- & -- & --
& -- & -- & --
& -- & -- & --
& -- & -- & --
& -- & -- & -- \\

HypCD~\cite{hypcd}
& 64.8 & 65.8 & 64.2
& \textbf{62.8} & 73.4 & 57.7
& 50.2 & 71.7 & 41.3
& 49.8 & 62.9 & 44.0
& 54.1 & 71.8 & 44.7
& 43.0 & 51.6 & 39.7
& 54.1 & 66.2 & 48.6 \\

\rowcolor{gray!20}
DCCL~\cite{dccl}
& 63.5 & 60.8 & 64.9
& 43.1 & 55.7 & 36.2
& 49.9 & 55.1 & 47.8
& 55.1 & 62.0 & 52.0
& 47.0 & 48.4 & 46.3
& 35.1 & 46.2 & 30.7
& 49.0 & 54.7 & 46.3 \\

\rowcolor{gray!20}
MGCE (Ours)
& \textbf{70.4} & 74.1 & \textbf{68.5}
& 61.5 & 75.9 & \textbf{54.5}
& \textbf{59.6} & 61.5 & \textbf{58.7}
& \textbf{58.2} & 68.6 & \textbf{53.6}
& \textbf{55.5} & 61.7 & \textbf{52.2}
& 45.3 & 54.5 & \textbf{41.7}
& \textbf{58.4} & 66.0 & \textbf{54.9} \\
\midrule
\multicolumn{22}{l}{(b) \textit{Comparison without the number of classes $K_U$ in the unlabeled dataset}} \\
\midrule
GCD~\cite{gcd}
& 51.1 & 56.4 & 48.4
& 39.1 & 58.6 & 29.7
& 51.9 & 65.6 & 46.2
& 49.6 & 65.1 & 42.7
& 50.8 & \textbf{65.3} & 43.1
& 40.8 & 49.0 & 37.6
& 47.2 & 60.0 & 41.3 \\

GPC~\cite{gpc}
& 52.0 & 55.5 & 47.5
& 38.2 & 58.9 & 27.4
& 44.2 & 51.8 & 41.0
& 40.4 & 59.5 & 32.0
& 43.3 & 62.3 & 33.2
& 37.3 & 50.9 & 31.9
& 42.6 & 56.5 & 35.5 \\

PIM~\cite{pim}
& 62.0 & \textbf{75.7} & 55.1
& 42.4 & 65.3 & 31.3
& 50.1 & \textbf{66.3} & 43.4
& 46.8 & 69.2 & 36.8
& 49.4 & 63.0 & 42.1
& 39.8 & 51.9 & 35.0
& 48.4 & 65.2 & 40.6 \\

\rowcolor{gray!20}
DCCL~\cite{dccl}
& 63.4 & 55.7 & 67.3
& 34.9 & 40.1 & 32.4
& 44.0 & 56.1 & 39.0
& 49.4 & 61.4 & 44.1
& 39.0 & 45.8 & 35.3
& 33.8 & 41.0 & 31.0
& 44.1 & 50.0 & 41.5 \\

\rowcolor{gray!20}
MGCE (Ours)
& \textbf{68.1} & 73.2 & \textbf{65.5}
& \textbf{57.8} & \textbf{72.7} & \textbf{50.7}
& \textbf{60.3} & 64.2 & \textbf{58.7}
& \textbf{55.8} & \textbf{72.4} & \textbf{48.4}
& \textbf{51.1} & 61.4 & \textbf{45.5}
& \textbf{44.0} & \textbf{55.1} & \textbf{39.6}
& \textbf{56.2} & \textbf{66.5} & \textbf{51.4} \\
\bottomrule
\end{tabular}
\end{table*}

\noindent\textbf{Evaluation Metric.} Following~\cite{gcd}, we report clustering accuracy (ACC), computed by aligning predicted cluster assignments with the ground-truth labels via the Hungarian algorithm~\cite{hungarian}.

\begin{table}[h]
\centering
\setlength{\tabcolsep}{1pt}
\renewcommand{\arraystretch}{0.6} 
\caption{Comparison on large-scale fine-grained datasets. The best results are highlighted in \textbf{bold}, and `--' indicates results not reported due to unavailable code.}

\label{tab:sota_2}
\footnotesize
\begin{tabular}{@{}l|ccc|ccc|ccc@{}}
\toprule
\multirow{2}{*}{Method} 
& \multicolumn{3}{c|}{Herbarium 19}
& \multicolumn{3}{c|}{NABirds}
& \multicolumn{3}{c}{Reptilia}\\ 
\cmidrule(lr){2-4} 
\cmidrule(lr){5-7} 
\cmidrule(lr){8-10} 

& All & Old & New 
& All & Old & New 
& All & Old & New  \\ 
\midrule
\multicolumn{10}{l}{(a) \textit{Comparison when $K_U$ is known}} \\
\midrule
GCD~\cite{gcd}
& 35.4 & 51.0 & 27.0  
& 35.4 & 67.7 & 21.1 
& 20.6 & 34.4 & 15.0  \\

XCon~\cite{xcon} 
& 38.1 & 58.3 & 27.3  
& 38.3 & 66.0 & 26.0 
& 23.8 & 43.0 & 16.1  \\

GPC~\cite{gpc}
& 36.5 & 51.7 & 27.9
& 38.5 & 72.0 & 23.7 
& 24.3 & 48.3 & 14.6 \\

InfoSieve~\cite{infosieve} 
& 41.0 & 55.4 & 33.2  
& 39.2 & 73.5 & 23.9
& 14.7 & 19.6 & 12.7  \\

SimGCD~\cite{simgcd} 
& 44.0 & 58.0 & 36.4  
& 35.7 & 76.1 & 17.7 
& 24.3 & 54.2 & 12.1  \\

PIM~\cite{pim} 
& 42.3 & 56.1 & 34.8  
& 36.2 & 73.1 & 19.8
& 26.6 & \textbf{57.5} & 14.0  \\

CMS~\cite{cms} 
& 36.4 & 54.9 & 26.4  
& 36.1 & 76.3 & 18.3 
& 22.9 & 48.0 & 12.8  \\

SPTNet~\cite{sptnet} 
& 43.4 & 58.7 & 35.2  
& 38.4 & \textbf{77.1} & 21.2 
& 25.8 & 49.8 & 16.2 \\

LegoGCD~\cite{legogcd} 
& 45.1 & 57.4 & 38.4  
& 37.4 & 75.8 & 20.2
& 24.4 & 49.9 & 14.1  \\

ProtoGCD~\cite{protogcd} 
& 44.5 & \textbf{59.4} & 36.5  
& 33.9 & 75.5 & 15.3 
& 25.8 & 52.6 & 14.9  \\

APL~\cite{aplgcd} 
& 44.9 & 58.1 & 37.9  
& -- & -- & -- 
& -- & -- & --  \\

HypCD~\cite{hypcd} 
& 40.5 & 57.0 & 31.6  
& 35.2 & 74.5 & 17.8 
& 30.3 & 49.4 & 22.7  \\

\rowcolor{gray!20} DCCL~\cite{dccl} 
& 32.0 & 40.3 & 27.5  
& 36.3 & 72.9 & 20.1 
& 22.8 & 40.9 & 15.5  \\

\rowcolor{gray!20} MGCE (Ours) 
& \textbf{45.3} & 53.2 & \textbf{41.1}  
& \textbf{49.7} & 73.2 & \textbf{39.3} 
& \textbf{30.5} & 45.7 & \textbf{24.4}  \\

\midrule
\multicolumn{10}{l}{(b) \textit{Comparison without $K_U$}} \\
\midrule
GCD~\cite{gcd} 
& 37.2 & 51.7 & 29.4  
& 34.6 & 68.1 & 19.8
& 22.3 & 35.8 & 16.8  \\

GPC~\cite{gpc} 
& 31.0 & 32.5 & 29.6  
& 37.0 & 71.7 & 21.6
& 22.8 & 47.7 & 12.8 \\

PIM~\cite{pim} 
& 42.0 & \textbf{55.5} & 34.7  
& 36.2 & 72.5 & 20.1
& 26.5 & \textbf{55.6} & 14.6 \\

\rowcolor{gray!20}
DCCL~\cite{dccl} 
& 30.5 & 37.6 & 26.7  
& 35.5 & 72.8 & 18.9  
& 22.5 & 40.5 & 15.2 \\

\rowcolor{gray!20}
MGCE (Ours)
& \textbf{43.1} & 53.6 & \textbf{37.4}  
& \textbf{53.3} & \textbf{74.7} & \textbf{43.8}  
& \textbf{29.8} & 50.6 & \textbf{21.4}  \\
\bottomrule
\end{tabular}
\end{table}

\noindent\textbf{Implementation Details.} 
We adopt the ViT-B/16 model pre-trained with DINO~\cite{dino} as the backbone. 
The representation of the [CLS] token is used as the image feature, and fine-tuning is applied only to the last transformer block of the backbone.  
The projection head $\phi(\cdot)$ and the parametric classifier $f_z$ follow the configuration of SimGCD~\cite{simgcd}, 
while the projection heads $\phi_C^r(\cdot)$ $(r=1,2,3)$ for concept experts are implemented as two-layer linear networks (768-2048-768). 
Training is conducted for 200 epochs with a batch size of 128, using an initial learning rate of 0.05 scheduled by cosine decay. For instance-level representation learning, following SimGCD~\cite{simgcd}, the balancing factor is set to $\lambda=0.35$, with temperature values $\tau_u=0.07$ and $\tau_l=1.0$. 
The classification objective uses $\tau_s=0.1$. 
When $K_U$ is known, the temperature $\tau_t$ in the unsupervised objective is initialized at 0.07 and annealed to 0.04 within the first 30 epochs, with $\epsilon=2$. 
For generic datasets, all parameters are kept identical except that $\lambda$ is reduced to 0.1. For concept-level representation learning, the mini-batch size is set to $\mathcal{B}_C=128$, with number of concepts $N^C=8$, number of instances per concept $N^I=16$, and temperature $\tau_c=0.05$. 
The threshold $\delta$ is set to 0.6 for fine-grained datasets and 0.5 for generic datasets. The scaling factor $\mathcal{R}$ is fixed at 0.6 and 
the loss weighting factor is $\alpha=0.1$. 
These two hyperparameters are chosen based on results from CUB and kept consistent across all datasets (see Sec.~\ref{sec:hyperparameter} for details). In dynamic concept generation, we employ FAISS~\cite{faiss} to accelerate the construction of similarity networks. 
All experiments are conducted on NVIDIA RTX 3090 GPUs, and reported results are averaged over three independent runs with different random seeds.

\subsection{Comparison with the State of the Art}
We compare the proposed MGCE framework with state-of-the-art approaches on nine challenging fine-grained datasets (Tab.~\ref{tab:sota_1} and Tab.~\ref{tab:sota_2}) and generic benchmarks (Tab.~\ref{tab:sota_3}). The competing methods include GCD~\cite{gcd}, XCon~\cite{xcon}, GPC~\cite{gpc}, InfoSieve~\cite{infosieve}, CMS~\cite{cms}, PIM~\cite{pim}, SimGCD~\cite{simgcd}, SPTNet~\cite{sptnet}, LegoGCD~\cite{legogcd}, ConGCD~\cite{congcd}, ProtoGCD~\cite{protogcd},  APL~\cite{aplgcd}, HypCD~\cite{hypcd}, and our baseline DCCL~\cite{dccl}. For a fair comparison, all methods adopt a DINO-pretrained ViT backbone~\cite{dino}. Since ConGCD, APL, and HypCD have multiple variants, we report the results of their variants that build upon the representation learning module of SimGCD. As some approaches assume prior knowledge of the number of novel classes $K_U$, we report results under two settings: (a) with the true number of novel classes $K_U$ provided, and (b) without knowledge of $K_U$.

\subsubsection{Results on Fine-Grained Datasets}  

\textbf{With known $K_U$.}  
Tab.~\ref{tab:sota_1} (a) and \ref{tab:sota_2} (a) show that our MGCE framework achieves state-of-the-art performance across both small- and large-scale fine-grained benchmarks. In particular, MGCE outperforms competing methods and obtains the best results on four of six small-scale datasets and all three large-scale datasets. Compared to the strongest parametric competitor Hyp-CD~\cite{hypcd}, MGCE yields an average improvement of 4.3\% in All ACC and 6.3\% in New ACC on small-scale datasets, and 6.5\% in All ACC and 10.9\% in New ACC on large-scale datasets. Against the strongest non-parametric competitor CMS~\cite{cms}, MGCE improves by 3.5\% in All ACC and 8.3\% in New ACC on small-scale datasets, and by 10.0\% in All ACC and 15.8\% in New ACC on large-scale datasets, highlighting the superior capability of MGCE in fine-grained category discovery. Compared with the GCD baseline~\cite{gcd}, our approach achieves an average gain of 11.8\% in All ACC and 15.0\% in New ACC across all fine-grained datasets. Furthermore, relative to GPC~\cite{gpc}, which also employs an iterative strategy of representation learning and category discovery, MGCE leverages multi-granularity experts to capture more comprehensive category information, achieving superior discovery accuracy. Specifically, MGCE surpasses GPC by an average of 11.3\% in All ACC and 15.2\% in New ACC across all fine-grained datasets. Notably, our MGCE framework with multi-granularity expert collaborative learning substantially enhances category understanding, resulting in an average improvement of 10.2\% in All ACC and 10.4\% in New ACC across all fine-grained datasets compared to our baseline DCCL~\cite{dccl}. The enhancement is particularly pronounced on large-scale datasets with more categories and higher difficulty, where MGCE achieves gains of 11.5\% in All ACC and 13.9\% in New ACC.

\noindent \textbf{With Unknown $K_U$.}
As shown in Tab.~\ref{tab:sota_1} (b) and \ref{tab:sota_2} (b), when $K_U$ is unknown, our MGCE model achieves the best performance across all fine-grained datasets. It exceeds the strong competitor PIM~\cite{pim} with average gains of 7.6\% in All ACC and 10.9\% in New ACC, and outperforms GPC~\cite{gpc} by 13.0\% in All ACC and 14.9\% in New ACC. These results highlight the strength of MGCE in discovering novel categories under realistic scenarios where the number of novel classes is unknown. Compared to DCCL~\cite{dccl}, our enhanced MGCE yields substantial improvements across all datasets, with average gains of 12.3\% in All ACC and 11.2\% in New ACC. It is also noteworthy that, although performance with unknown $K_U$ is generally lower than with known $K_U$, MGCE achieves higher accuracy under the unknown-$K_U$ setting on the Animalia and NABirds datasets.

\subsubsection{Results on Generic Datasets} 
We compare our MGCE framework with two baselines, GCD~\cite{gcd} and SimGCD~\cite{simgcd}, as well as state-of-the-art methods, ProtoGCD~\cite{protogcd}, ConGCD~\cite{congcd}, and HypCD~\cite{hypcd}, on two generic benchmarks, as shown in Tab.~\ref{tab:sota_3}. While MGCE is primarily designed for fine-grained scenarios that involve rich hierarchical relationships, the improvements on generic benchmarks are more modest, reflecting their relatively limited hierarchical structure. Nonetheless, MGCE achieves competitive performance on generic datasets. Compared with the GCD and Sim-GCD baselines, MGCE achieves average improvements of 9.7\% over GCD and 1.7\% over SimGCD in All ACC across the two datasets. Relative to the strongest competitor HypCD~\cite{hypcd}, MGCE is 1.2\% lower in All ACC on average, but it is worth noting that MGCE outperforms HypCD by 0.6\% in Old ACC. Furthermore, compared with our baseline DCCL~\cite{dccl}, MGCE achieves average gains of 5.4\% in All ACC and 5.7\% in New ACC, highlighting that the proposed multi-granularity expert collaborative learning strategy can also effectively improve category discovery accuracy in generic scenarios.

\begin{table}[t]
\centering
\caption{Comparison with baselines and state-of-the-art methods on generic datasets. The best results are shown in \textbf{bold}.}

\setlength{\tabcolsep}{4pt}
\renewcommand{\arraystretch}{0.6} 
\label{tab:sota_3}
\footnotesize
\begin{tabular}{@{}l|ccc|ccc@{}}
\toprule
\multirow{2}{*}{Method} 
& \multicolumn{3}{c|}{Cifar100}
& \multicolumn{3}{c}{ImageNet100}\\ 
\cmidrule(lr){2-4} 
\cmidrule(lr){5-7} 

& All & Old & New 
& All & Old & New  \\ 
\midrule
GCD~\cite{gcd} & 73.0 & 76.2 & 66.5 & 74.1 & 89.8 & 66.3 \\

SimGCD~\cite{simgcd} & 80.1 & 81.2 & 77.8 & 83.0 & 93.1 & 77.9 \\

ProtoGCD~\cite{protogcd} & 81.9 & 82.9 & 80.0 & 84.0 & 92.2 & 79.9\\

ConGCD~\cite{congcd} & 81.3 & 82.5 & 78.9 & 83.5 & 92.8 & 78.6\\

HypCD~\cite{hypcd} & \textbf{82.4} & 83.1 & \textbf{81.2} & \textbf{86.5} & 93.7 & \textbf{83.0}\\

\rowcolor{gray!20}
DCCL~\cite{dccl} &75.3 & 76.8 & 70.2 & 80.5 & 90.5 & 76.2 \\
\rowcolor{gray!20}
MGCE (Ours)& 82.2 & \textbf{83.6} & 79.5 & 84.3 & \textbf{94.4} & 78.2\\
\bottomrule
\end{tabular}
\end{table}

\begin{table*}[!t]
\centering
\caption{
    Ablation study on the proposed components. 
    \textit{CRL} refers to Concept-Level Representation Learning, 
    \textit{IRL} to Instance-Level Representation Learning, 
    \textit{ME} to Multi-Expert Conceptual Learning, 
    and \textit{CL} to Collaborative Learning among Multiple Experts.
}
\label{tab:ablation_study}
\setlength{\tabcolsep}{2pt}
\begin{tabular}{@{}ccccc|ccc|ccc|ccc|ccc|ccc@{}}
\toprule
\multirow{2}{*}{Configuration} & \multirow{2}{*}{\textit{CRL}} & \multirow{2}{*}{\textit{IRL}} & \multirow{2}{*}{\textit{ME}} & \multirow{2}{*}{\textit{CL}} & \multicolumn{3}{c|}{CUB} & \multicolumn{3}{c|}{Stanford Cars} & \multicolumn{3}{c|}{Fungi} & \multicolumn{3}{c|}{Herbarium 19} & \multicolumn{3}{c}{NABirds} \\
\cmidrule(lr){6-8} 
\cmidrule(lr){9-11}
\cmidrule(lr){12-14}
\cmidrule(lr){15-17}
\cmidrule(lr){18-20}
 &  &  & &  & All & Old & New & All & Old & New & All & Old & New & All & Old & New & All & Old & New \\
\midrule
DCCL    
& \textcolor{mydarkgreen}{\checkmark} 
& \textcolor{red}{\textbf{\sffamily x}} 
& \textcolor{red}{\textbf{\sffamily x}} 
&\textcolor{red}{\textbf{\sffamily x}} 
& 63.5 & 60.8 & 64.9 
& 43.1 & 55.7 & 36.2 
& 55.1 & 62.0 & 52.0 
& 32.0 & 40.3 & 27.5 
& 36.3 & 72.9 & 20.1 \\

\midrule

\uppercase\expandafter{\romannumeral1}  
& \textcolor{red}{\textbf{\sffamily x}} 
& \textcolor{mydarkgreen}{\checkmark} 
& \textcolor{red}{\textbf{\sffamily x}} 
& \textcolor{red}{\textbf{\sffamily x}} 
& 63.2 & 69.2 & 60.2
& 56.2 & 70.0 & 49.5
& 54.5 & 69.0 & 48.1
& 43.1 & 52.7 & 32.4
& 39.5 & 63.9 & 29.1\\

\uppercase\expandafter{\romannumeral2}  
& \textcolor{mydarkgreen}{\checkmark} 
& \textcolor{mydarkgreen}{\checkmark}
& \textcolor{red}{\textbf{\sffamily x}} 
& \textcolor{red}{\textbf{\sffamily x}} 
& 67.0 & 71.3 & 64.9
& 57.1 & 75.1 & 48.5
& 56.0 & 66.0 & 51.5
& 43.4 & 50.3 & 39.7
& 44.0 & \textbf{75.5} & 30.1\\

\uppercase\expandafter{\romannumeral3}  
& \textcolor{mydarkgreen}{\checkmark} 
& \textcolor{mydarkgreen}{\checkmark}
& \textcolor{mydarkgreen}{\checkmark} 
& \textcolor{red}{\textbf{\sffamily x}} 
& 69.0 & 72.6 & 67.2
& 60.4 & 75.7 & 53.1
& 57.9 & \textbf{68.6} & 53.2 
& 44.4 & 52.5 & 40.1
& 49.0 & 75.1 & 37.4\\

\rowcolor{gray!20}  \uppercase\expandafter{\romannumeral4} 
& \textcolor{mydarkgreen}{\checkmark} 
& \textcolor{mydarkgreen}{\checkmark} 
& \textcolor{mydarkgreen}{\checkmark} 
& \textcolor{mydarkgreen}{\checkmark} 
& \textbf{70.4} & \textbf{74.1} & \textbf{68.5} 
& \textbf{61.5} & \textbf{75.9} & \textbf{54.5} 
& \textbf{58.2} & \textbf{68.6} & \textbf{53.6} 
& \textbf{45.3} & \textbf{53.2} & \textbf{41.1} 
& \textbf{49.7} & 73.2 & \textbf{39.3} \\
\bottomrule
\end{tabular}
\end{table*}

\subsection{Ablation Study}
We conduct an ablation study to assess the contribution of different components within our MGCE framework across five fine-grained benchmarks, including three small-scale datasets (CUB, Stanford Cars, and Fungi) and two large-scale datasets (Herbarium19 and NABirds). Specifically, we design two sets of experiments. \textbf{(i) Com-ponent-wise Evaluation:} We start from the DCCL baseline and progressively incorporate the Concept-Level Representation Learning (CRL), the Instance-Level Representation Learning (IRL), the Multi-Expert Conceptual Learning (ME), and the Collaborative Learning among Multiple Experts (CL) to evaluate the effect of each component. \textbf{(ii) Single vs. Multi-Expert Strategy:} We construct a single conceptual expert by assigning different neighborhood sizes $k_{\text{nn}}^r$ ($r=1,2,3$) to represent varying concept granularities, and then compare these variants with the full MGCE to evaluate the effectiveness of the Multi-Granularity Experts Collaborative Learning strategy.

\noindent \textbf{Component-wise Evaluation.} 
As shown in Tab.~\ref{tab:ablation_study}, DCCL employs only the concept-level representation learning loss, while Variant~\uppercase\expandafter{\romannumeral1} uses only the instance-level representation learning loss. 
Variant~\uppercase\expandafter{\romannumeral2}, which combines both concept-level and instance-level representation learning, achieves consistently better results, surpassing DCCL by 7.5\% and Variant~\uppercase\expandafter{\romannumeral1} by 2.2\% in terms of All ACC. 
These results demonstrate that concept-level and instance-level representation learning improve the model’s category and concept understanding from complementary perspectives. 
Building on Variant~\uppercase\expandafter{\romannumeral2}, 
Variant~\uppercase\expandafter{\romannumeral3} and 
Variant~\uppercase\expandafter{\romannumeral4} progressively introduce 
Multi-Expert Conceptual Learning (ME) and Collaborative Learning among Multiple Experts (CL). 
The introduction of ME yields average improvements of 2.6\% and 3.3\% in All ACC and New ACC, 
showing that learning concept information at multiple granularities enhances the model’s conceptual understanding and leads to more accurate recognition. 
Further incorporating CL brings consistent improvements across all datasets, 
with average gains of 0.9\% and 1.2\% in All ACC and New ACC, indicating that knowledge sharing among experts enables more precise category division.

\begin{figure}[h]
  \begin{center}
\includegraphics[width=0.49\textwidth]{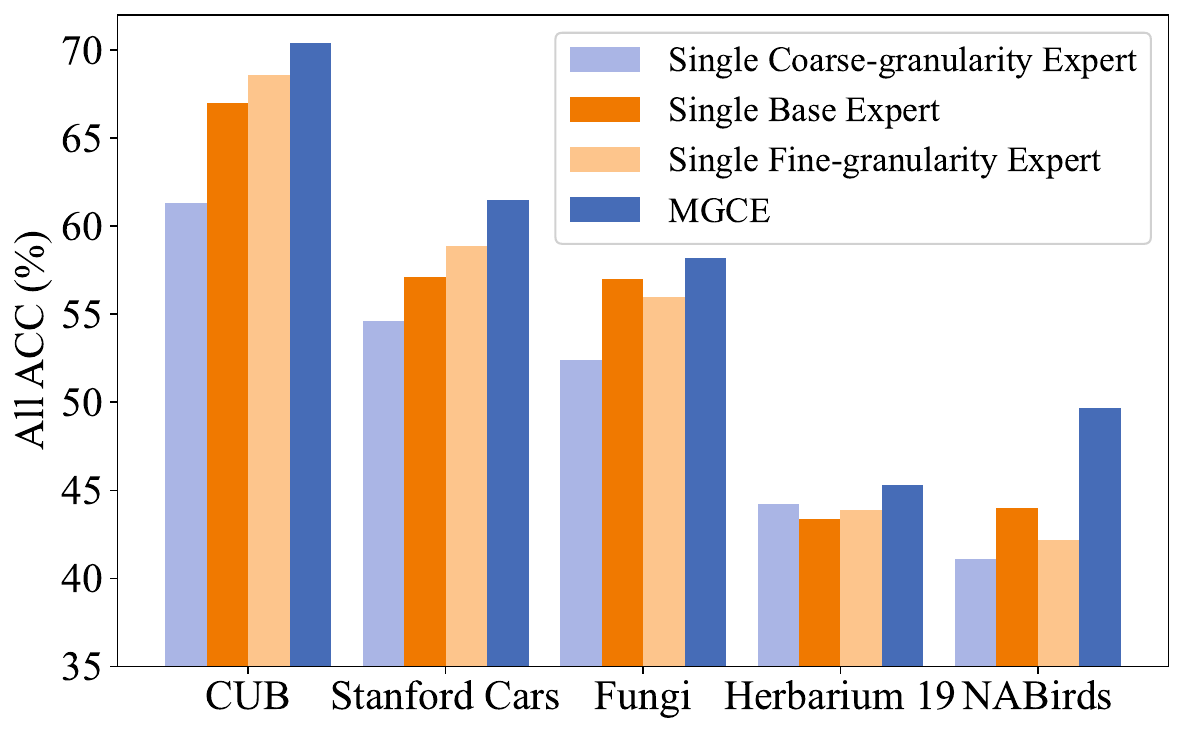}
\caption{Comparison of All ACC between multi-expert and single-expert learning across five datasets.
\label{fig:ablation_allacc}}

  \end{center}
\end{figure}

\begin{figure}[t]
  \begin{center}
\includegraphics[width=0.49\textwidth]{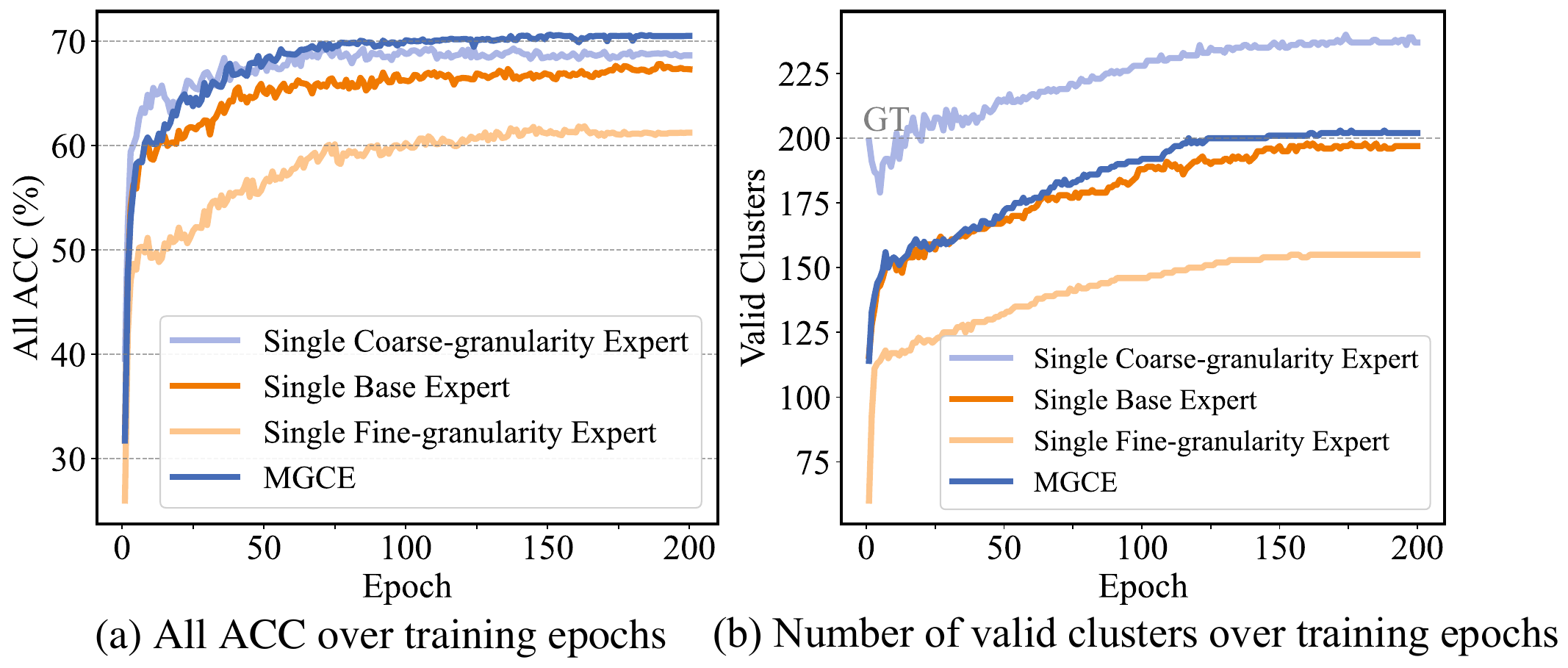}
\caption{Evolution of All ACC and the number of effective categories during training on CUB, for single-expert and multi-expert learning.
\label{fig:ablation_cubtraining}}

  \end{center}
\end{figure}

\noindent \textbf{Single vs. Multi-Expert Strategy.} 
To further analyze the effectiveness of the proposed Multi-Granularity Experts Collaborative Learning (MECL), 
we construct a single conceptual expert by assigning different neighborhood sizes $k_{\text{nn}}^r$ ($r=1,2,3$) to represent varying concept granularities. 
Fig.~\ref{fig:ablation_allacc} shows the comparison of All ACC across five datasets, 
while Fig.~\ref{fig:ablation_cubtraining} depicts the training dynamics on the CUB dataset, 
including the evolution of All ACC and the number of valid clusters. 
A fine-grained expert employs a smaller neighborhood radius when building the similarity graph, 
resulting in fewer but more precise connections, which leads to finer category partitions. 
As shown in Fig.~\ref{fig:ablation_cubtraining}, this enables the discovery of a larger number of valid categories. 
In contrast, coarse-grained experts use larger neighborhoods and thus capture broader category structures.  
Overall, the results in Fig.~\ref{fig:ablation_allacc} demonstrate that learning from multiple granularities, 
together with knowledge transfer among experts, consistently improves category discovery accuracy across all datasets.

\subsection{Evaluation}
\noindent \textbf{Evaluation on the Estimated Number of Clusters.} 
Our MGCE framework is applicable under both known and unknown numbers of classes. 
When the number of classes is unknown, we directly apply the Semi-Infomap algorithm on the learned feature representations to obtain both the category partitions and an estimate of the class count. 
As shown in Tab.~\ref{tab:k}, we compare MGCE with state-of-the-art non-parametric methods for class number estimation. 
Overall, MGCE achieves the lowest error rate on 4 out of 9 datasets, with an average error rate of 25\% across all datasets, reducing the error by 3.3\% compared with GCD. 
Furthermore, GCD relies on a two-stage pipeline, where category estimation is performed via Semi-$k$-means after feature learning, decoupled from the model’s optimization objective. 
In contrast, MGCE unifies category estimation with concept-level contrastive learning and multi-expert collaboration, enabling a dynamic and end-to-end framework that benefits directly from the estimation process. 
Compared with GPC, MGCE reduces the average error rate by 14.5\%, highlighting its advantage in mitigating error accumulation during the iterative process of representation learning and category discovery. 
Compared with the baseline DCCL, MGCE achieves a 26.6\% lower error rate. 
This result underscores the effectiveness of our proposed Multi-Granularity Experts Collaborative Learning in more accurately identifying and partitioning novel categories, thereby enhancing representation learning.

\begin{table}[h]
\centering
\setlength{\tabcolsep}{0.8pt}
\small
\vspace{-0.5cm}
\caption{Comparison of estimated class numbers (K) and estimation error rates (Err\%).}
\label{tab:k}
\begin{tabular}{l|c|c|c|c|c}
\toprule
Dataset & \textit{GT} & GCD~\cite{gcd} & GPC~\cite{gpc} & \cellcolor{gray!20} DCCL~\cite{dccl} & \cellcolor{gray!20} MGCE \\
\midrule
CUB            & 200 & 231/15.5 & 212/6.0   & \cellcolor{gray!20} 172/14.0   & \cellcolor{gray!20} \textbf{202/1.0} \\
Stanford Cars  & 196 & 230/17.3 & 201/2.6   & \cellcolor{gray!20} \textbf{192/2.0}    & \cellcolor{gray!20} 203/3.6 \\
Animalia       & 77  & 40/48.1  & 115/49.4  & \cellcolor{gray!20} 177/129.9  & \cellcolor{gray!20} \textbf{53/31.2} \\
Fungi          & 121 & 86/28.9  & 181/49.6  & \cellcolor{gray!20} 184/52.1   & \cellcolor{gray!20} \textbf{112/7.4} \\
Mollusca       & 93  & \textbf{54/41.9}  & 139/49.5  & \cellcolor{gray!20} 194/108.6  & \cellcolor{gray!20} 43/53.8 \\
Actinopterygii & 53  & 42/20.8  & 79/49.1   & \cellcolor{gray!20} 77/45.3    & \cellcolor{gray!20} \textbf{61/15.1} \\
Herbarium 19   & 683 & \textbf{520/23.8} &   1025/50.0        & \cellcolor{gray!20} 421/38.4   & \cellcolor{gray!20} 459/32.8 \\
NABirds        & 555 & \textbf{503/9.4}  & 832/49.9  & \cellcolor{gray!20} 644/16.0    & \cellcolor{gray!20} 715/28.8 \\
Reptilia       & 289 & \textbf{148/48.8} & 433/49.8  & \cellcolor{gray!20} 120/58.5   & \cellcolor{gray!20} 140/51.6 \\
\bottomrule
\end{tabular}
\vspace{-0.5cm}
\end{table}

\noindent \textbf{Evaluation of Projection Heads.} 
During training, our MGCE framework employs projection heads $\phi_C^r$ ($r=1,2,3$) to map image features for concept-level contrastive learning, rather than directly using backbone features. 
To evaluate this design choice, we compare different projection head configurations, as shown in Tab.~\ref{tab:projection}. 
The variant ``No Projection Head'' applies Semi-Infomap and concept-level contrastive learning directly to the backbone feature representations.  
Among the tested settings, the 2-layer projection head achieves the best performance, improving All ACC by 1.8\% over the 1-layer head and by 1.1\% over the 3-layer head. 
On average, introducing projection heads yields superior results in both All ACC and New ACC. 
This improvement can be attributed to alleviating feature entanglement in the shared projection space and reducing interference among experts.

\begin{table}[h]
\centering
\setlength{\tabcolsep}{4pt}
\caption{Comparison of projection head configurations: different depths (1-layer, 2-layer, and 3-layer) and the case without a projection head.}
\label{tab:projection}
\begin{tabular}{@{}ccccccc@{}}
\toprule
\multirow{2}{*}{Configuration} & \multicolumn{3}{c}{CUB} & \multicolumn{3}{c}{Stanford Cars} \\
\cmidrule(lr){2-4} 
\cmidrule(lr){5-7} 
 & All & Old & New & All & Old & New \\
\midrule
No Projection Head & 67.9 & 72.0 & 65.8 & 60.5 & \textbf{76.5} & 52.8  \\
1-layer Head       & 69.4  & \textbf{75.6} & 66.2 & 58.9 & 74.1 & 51.5 \\
\rowcolor{gray!20} 2-layer Head & \textbf{70.4} & 74.1 & \textbf{68.5}
& \textbf{61.5} & 75.9 & \textbf{54.5}   \\
3-layer Head       & 69.5 & 72.0 & 68.2 & 60.3 & 75.3 &  53.0 \\
\bottomrule
\end{tabular}
\vspace{-1cm}
\end{table}

\subsection{Hyperparameter Analysis}
\label{sec:hyperparameter}
Our MGCE framework involves two key hyperparameters: the loss weighting factor $\alpha$ and the scaling factor $\mathcal{R}$. Both are tuned on the CUB dataset. This choice may not yield optimal results on other datasets; however, to avoid overfitting to individual datasets and to ensure fair cross-dataset evaluation, we fix the same hyperparameters across all fine-grained benchmarks. To further assess their robustness and impact, we conduct a detailed sensitivity analysis of these hyperparameters on six fine-grained datasets.

\begin{figure}[h]
  \begin{center}
\includegraphics[width=0.49\textwidth]{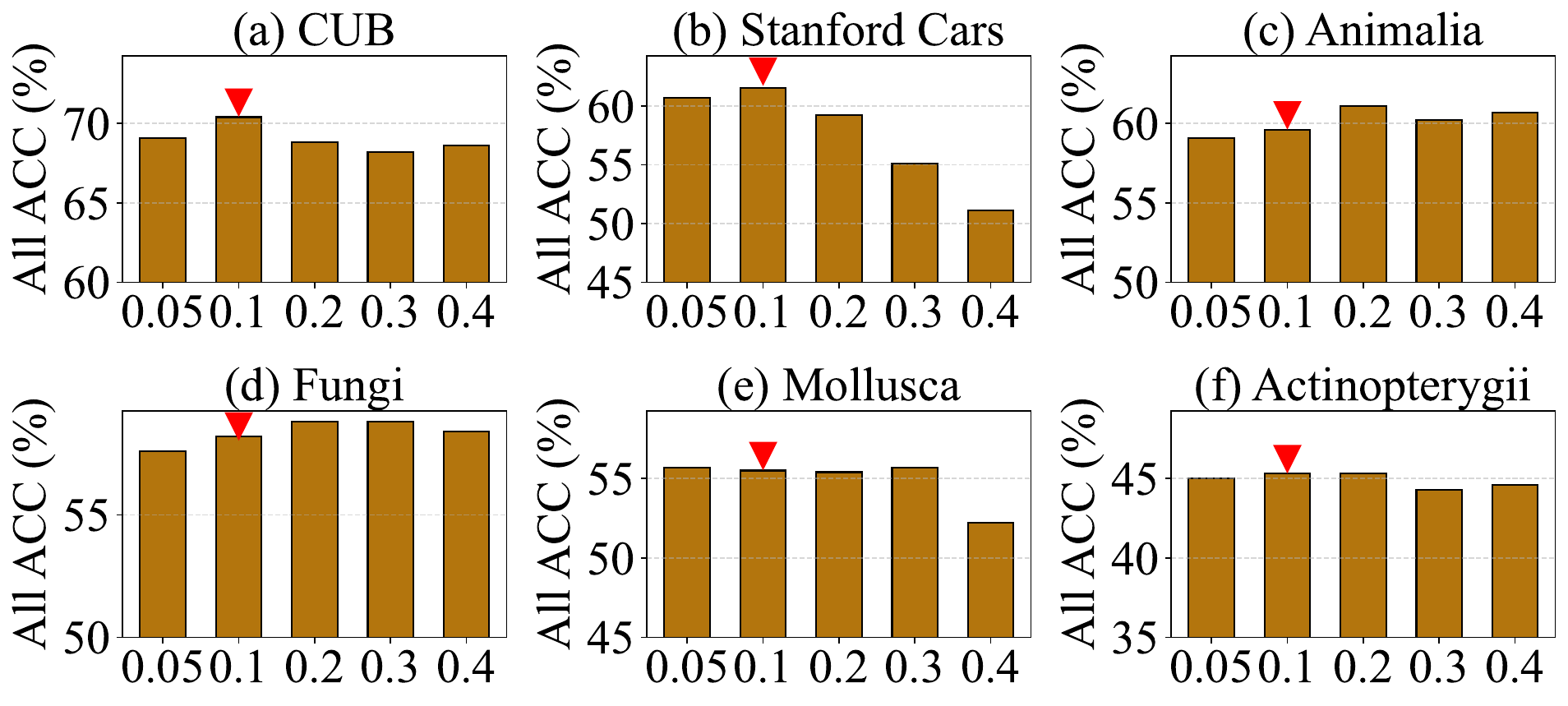}
\caption{Sensitivity of performance to the weighting factor $\alpha$ across six datasets.
\label{fig:hyperparam_1}}
  \end{center}
\vspace{-1cm}
\end{figure}

\noindent \textbf{Weighting Factor $\alpha$.} 
We analyze the effect of the weighting factor $\alpha$ in Eq.~\ref{eq:total_loss} on six datasets, as illustrated in Fig.~\ref{fig:hyperparam_1}, and report results in terms of All ACC. 
Overall, $\alpha$ shows a relatively stable influence on performance, with smaller values generally leading to better results. 
We therefore set $\alpha=0.1$ as the default value, based on the results on the CUB dataset.

\begin{figure}[h]
  \begin{center}
\includegraphics[width=0.49\textwidth]{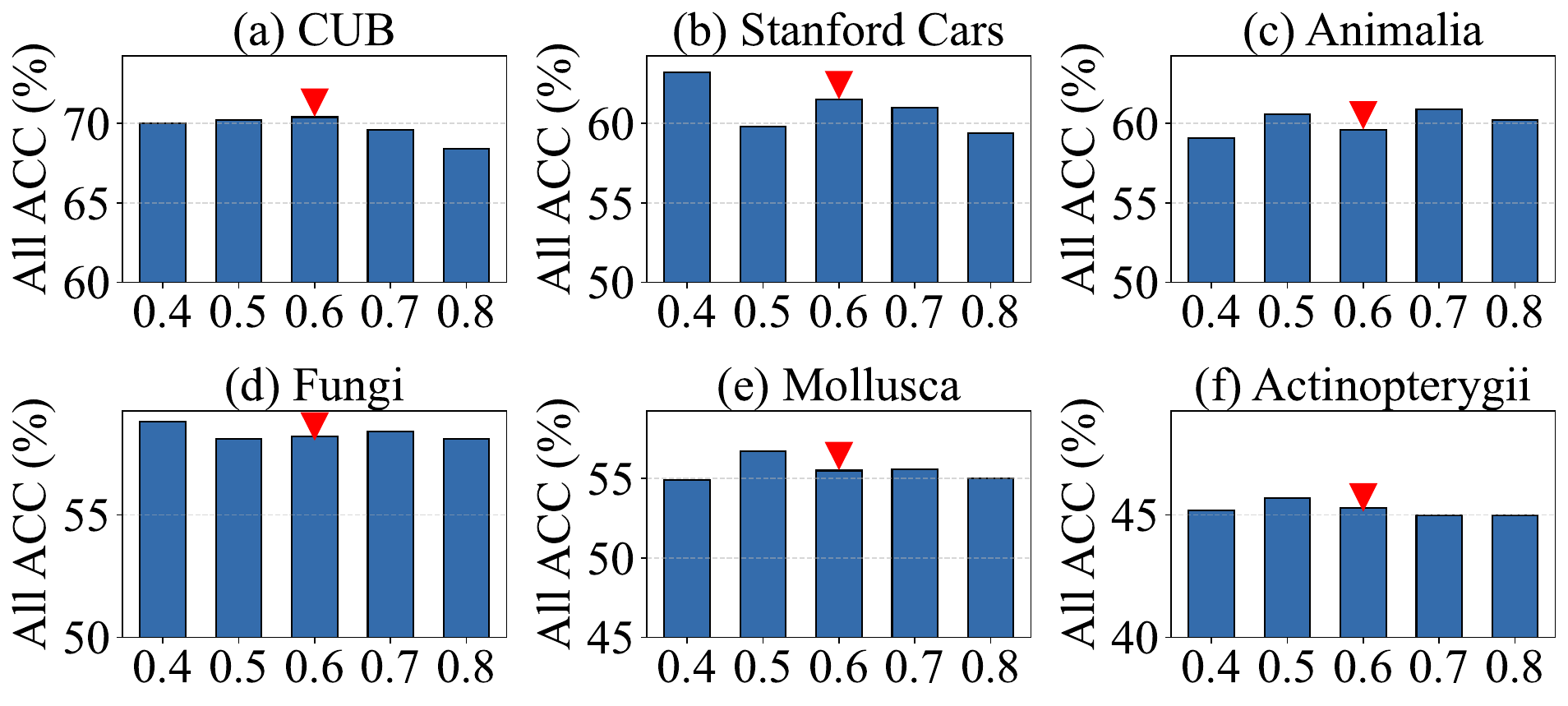}
\caption{Sensitivity of performance to the scaling factor $\mathcal{R}$ across six datasets.
\label{fig:hyperparam_2}}
  \end{center}
\vspace{-1cm}
\end{figure}

\noindent \textbf{Scaling Factor $\mathcal{R}$.} 
The scaling factor $\mathcal{R}$ in Eq.~\ref{eq:scaling} controls the extent of granularity variation among experts. 
Specifically, smaller values of $\mathcal{R}$ lead to greater differences in information granularity across the three experts, while larger values make the experts more similar in granularity. 
We evaluate the effect of $\mathcal{R}$ on six datasets, as illustrated in Fig.~\ref{fig:hyperparam_2}, and report results in terms of All ACC. 
Overall, the results indicate that values of $\mathcal{R}$ in the range 0.4--0.6 achieve the best performance. 
We therefore set $\mathcal{R}=0.6$ as the default value, based on the results on the CUB dataset.

\begin{figure*}[h!]
  \begin{center}
\includegraphics[width=0.98\textwidth]{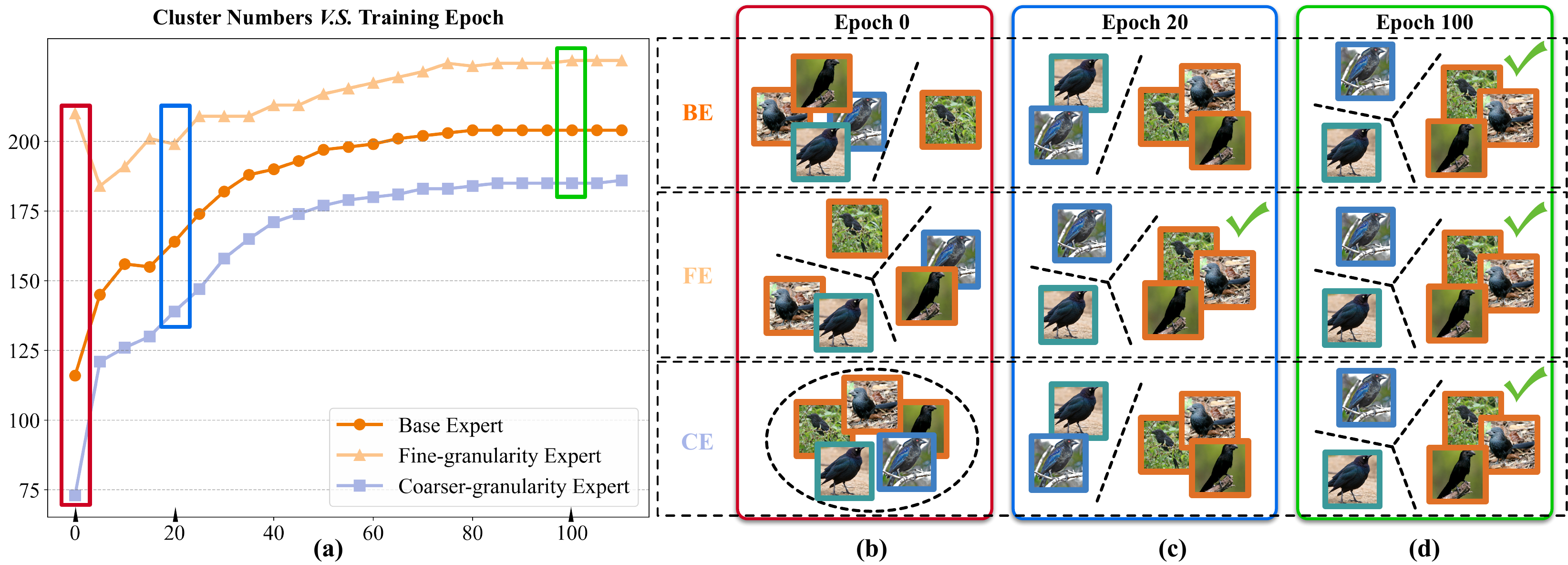}
\caption{Case analysis on the CUB dataset. Subfigure (a) shows the number of categories identified by the three experts in MGCE across training epochs. Subfigures (b), (c), and (d) illustrate the clustering results of the three experts at the 0-th, 20-th, and 100-th epochs, respectively, for three fine-grained bird species. Images with borders of the same color belong to the same category.}
\vspace{-0.8cm}
\label{fig:case}
  \end{center}
\end{figure*}

\subsection{Further Analysis}
\label{exp:further_analysis}
\noindent \textbf{Case Study.} 
We present a case study on the CUB dataset to illustrate the collaborative learning process among multi-granularity experts. 
We focus on three visually similar bird species: \textit{Groove-billed Ani}, \textit{Bronzed Cowbird}, and \textit{Brewer’s Blackbird}. 
As illustrated in Fig.~\ref{fig:case} (a), the effective number of categories estimated by the three experts is shown during training, while Figs.~\ref{fig:case} (b)--(d) display the category partitions at epochs 0, 20, and 100, respectively. In the initial epoch (red box in Fig.~\ref{fig:case} (a)), significant differences in the number of predicted categories are observed among the experts. Specifically, as shown in Fig.~\ref{fig:case} (b), the coarse-grained expert groups the three visually similar, predominantly black birds into a single category. In contrast, the fine-grained experts identify more distinct categories by considering not only color but also subtle morphological cues, such as grouping images of \textit{Groove-billed Ani} and \textit{Brewer’s Blackbird} based on similar postures. As training progresses, collaborative learning facilitates the exchange of category insights among experts. This leads to a more consistent trend in the number of predicted categories. By the 20th epoch (blue box in Fig.~\ref{fig:case} (a)), all three experts converge in their judgments for these categories, although \textit{Brewer’s Blackbird} and \textit{Bronzed Cowbird} remain grouped together, as shown in Fig.~\ref{fig:case} (c). By the 100th epoch (green box in Fig.~\ref{fig:case} (a)), the experts have unified their insights and accurately distinguished all three fine-grained categories, as shown in Fig.~\ref{fig:case} (d).

\noindent \textbf{Exploration of Multi-Expert Learning Strategies.} 
Our MGCE framework employs a multi-expert learning strategy by integrating fine-grained, coarse-grained, and basic experts. 
We first assess the contributions of the fine-grained and coarse-grained experts by systematically removing them individually. 
As shown in Tab.~\ref{tab:expert_strategy}, removing the coarse-grained expert leads to an average drop of 1.4\% in All ACC across three datasets, while removing the fine-grained expert results in a larger drop of 2.7\%. 
These results highlight that complementary signals from different granularities enhance the framework’s effectiveness in representing and partitioning categories.  
We further investigate the effect of varying the number of experts by adding additional ones with a scaling factor of $\mathcal{R}=0.4$. 
As shown in Tab.~\ref{tab:expert_strategy}, adding more experts beyond the current configuration does not provide further benefits: introducing extra fine-grained experts decreases All ACC by 0.5\% on average, while adding coarse-grained experts decreases it by 0.7\%. 
This demonstrates that the current multi-expert setup strikes a suitable balance between granularity and performance.

\begin{table}[t]
\centering
\caption{Evaluation of different expert configurations, where ``BE'' denotes Base Expert, ``CE'' denotes Coarse-granularity Expert, and ``FE'' denotes Fine-granularity Expert.}
\vspace{0.2cm}
\label{tab:expert_strategy}
\setlength{\tabcolsep}{1pt}
\begin{tabular}{@{}cccccccccc@{}}
\toprule
\multirow{2}{*}{Method} 
& \multicolumn{3}{c}{CUB} 
& \multicolumn{3}{c}{Animalia} 
& \multicolumn{3}{c}{Mollusca} \\
\cmidrule(lr){2-4} 
\cmidrule(lr){5-7} 
\cmidrule(lr){8-10} 
 & All & Old & New & All & Old & New & All & Old & New \\
\midrule
Single BE
& 67.0 & 71.3 & 64.9
& 53.6 & 64.3 & 49.2
& 52.2 & 55.6 & 50.4 \\

BE+CE 
& 66.9 & 71.5 & 64.6
& 55.9 & \textbf{66.8} & 51.3
& 54.7 & 60.8 & 51.5\\

BE+FE
& 70.0 & 74.2 & 67.9
& 57.9 & 65.6 & 54.6  
& 53.4 & 55.9 & 52.1 \\
\midrule

MGCE+CE
& 69.4 & 74.8 & 66.6
& 58.7 & 63.0 & 56.9
& 55.2 & \textbf{63.9} & 50.5 \\

MGCE+FE
& \textbf{70.5} & \textbf{75.5} &  68.0
& 58.5 & 62.2 & 57.0
& 55.0 & 60.1 & 52.0 \\
\midrule
\rowcolor{gray!20} MGCE (Ours)
& 70.4 & 74.1 & \textbf{68.5}
& \textbf{59.6} & 61.5 & \textbf{58.7}
& \textbf{55.5} & 61.7 & \textbf{52.2} \\
\bottomrule
\end{tabular}
\vspace{-0.5cm}
\end{table}

\subsection{Limitation and Discussion.}
Our MGCE achieves state-of-the-art performance on fine-grained datasets and significantly improves upon the DCCL baseline, primarily due to its ability to exploit the rich multi-granularity information inherent in fine-grained data. However, since MGCE relies on dynamically discovered multi-granularity information, its effectiveness diminishes in scenarios where such hierarchical knowledge is limited or where classes exhibit large intra-class variation. As a result, the improvements in generic datasets are more modest compared to fine-grained scenarios.

\section{Conclusion}
In this paper, we seek to address the GCD challenges from the perspective of dynamic extraction and multi-granular exploitation of concept-level knowledge contained in labeled and unlabeled data. To achieve this goal, we propose a Multi-Granularity Conceptual Experts (MGCE) framework consisting of two complementary modules: Dynamic Conceptual Contrastive Learning (DCCL) and Multi-Granularity Experts Collaborative Learning (MECL). DCCL focuses on dynamically generating conceptual knowledge tailored for expert learning across different training epochs, while MECL emphasizes the integration of multi-granularity experts to perform collaborative learning within each epoch. This novel paradigm adaptively injects complementary concept-level knowledge into the training process, enabling the model to learn more discriminative representations for accurate category discovery. Extensive experimental results demonstrate that our MGCE achieves new state-of-the-art performance on GCD tasks.


\bibliographystyle{spmpsci}      
\bibliography{egbib}   


\end{document}